\documentclass{article}

\usepackage[preprint]{ms}

\usepackage{natbib} 
\usepackage[utf8]{inputenc} 
\usepackage[T1]{fontenc}    
\usepackage{hyperref}       
\usepackage{url}            
\usepackage{booktabs}       
\usepackage{amsfonts}       
\usepackage{nicefrac}       
\usepackage{microtype}      
\usepackage{colortbl}
\usepackage{color,soul}

\usepackage{amsmath}
\usepackage{amssymb}
\usepackage{bbm}
\usepackage{array}
\usepackage{mathtools}
\usepackage{amsthm}
\usepackage{amsfonts}
\usepackage{centernot}
\usepackage{xspace}

\newcommand{\x}{\mathbf{x}}

\usepackage{xspace}

\usepackage[capitalize]{cleveref}

\usepackage[dvipsnames]{xcolor}
\usepackage{caption}
\usepackage{subcaption}
\usepackage{multirow}
\usepackage{enumitem}
\usepackage{wrapfig}
\usepackage{pifont}
\newcommand{\cmark}{\textcolor{green!60!black}{\ding{51}}} 
\newcommand{\xmark}{\textcolor{red}{\ding{55}}}            
\colorlet{LightCyan}{cyan!10!white}
\colorlet{bluecorner}{RoyalBlue}
\colorlet{bluefill}{Cyan!10}
\colorlet{yellowfill}{yellow!30}
\colorlet{redfill}{red!10}
\newcommand*\circled[1]{\tikz[baseline=(char.base)]{
    \node[shape=circle,draw,inner sep=0.5pt] (char) {\sffamily\footnotesize #1};}}

\usepackage{tikz}
\usetikzlibrary{shapes,positioning,fit,decorations.pathreplacing,patterns,calligraphy,calc,arrows}
\usepackage[skins]{tcolorbox}

\usepackage{mathtools}

\makeatletter
\newcommand{\colormathbox}[3][\mathord]{%
  #1{%
    \setlength{\fboxsep}{0pt}%
    \mathpalette\color@mathbox{{#2}{#3}}%
  }%
}
\newcommand{\color@mathbox}[2]{%
  \color@@mathbox#1#2%
}
\newcommand{\color@@mathbox}[3]{%
  \colorbox{#2}{$#1\m@th#3$}%
}
\makeatother

\newcommand{\colorboxtight}[2]{\setlength{\fboxsep}{0.75pt}\colorbox{#1}{#2}}

\title{Object-level Self-Distillation for Vision Pretraining}

\author{%
    \c{C}a\u{g}lar H{\i}zl{\i} \\
    Aalto University\\
    \texttt{caglar.hizli@aalto.fi} \\
    \And
    \c{C}a\u{g}atay Y{\i}ld{\i}z \\
    University of Tübingen \\
    Tübingen AI Center \\
    \And
    Pekka Marttinen \\
    Aalto University\\
}

\begin{document}

\maketitle

\begin{abstract}
    State-of-the-art vision pretraining methods rely on image‑level self‑distillation from object‑centric datasets such as ImageNet, implicitly assuming each image contains a single object. This assumption does not always hold: many ImageNet images already contain multiple objects. Further, it limits scalability to scene‑centric datasets that better mirror real‑world complexity. We address these challenges by introducing \textbf{O}bject-level Self-\textbf{Dis}tillation (ODIS), a pretraining approach that shifts the self-distillation granularity from whole images to individual objects.
    Using object-aware cropping and masked attention, ODIS isolates object-specific regions, guiding the transformer toward semantically meaningful content and transforming a noisy, scene-level task into simpler object-level sub-tasks. We show that this approach improves visual representations both at the image and patch levels. Using masks at inference time, our method achieves an impressive $82.6\%$ $k$-NN accuracy on ImageNet1k with ViT-Large.
\end{abstract}

\section{Introduction}

Vision Transformers (ViTs) \citep{dosovitskiy2020image} have emerged as foundation models for diverse visual tasks--from unsupervised segmentation to dense correspondence and appearance transfer--\citep{amir2021deep,tumanyan2022splicing,ofri2023neural,hamilton2022unsupervised}, as their
frozen features capture rich, transferable semantic information. Like large language models, ViTs derive much of their representational power from large-scale self-supervised pretraining \citep{caron2021emerging,zhou2021ibot,oquab2023dinov2}. State-of-the-art pretraining methods typically employ a teacher-student architecture \citep{tarvainen2017mean} and self-distillation \citep{caron2021emerging}. In these methods, the teacher network provides reference embeddings, guiding the student to align its representations at a chosen granularity--most often at the image and patch level.

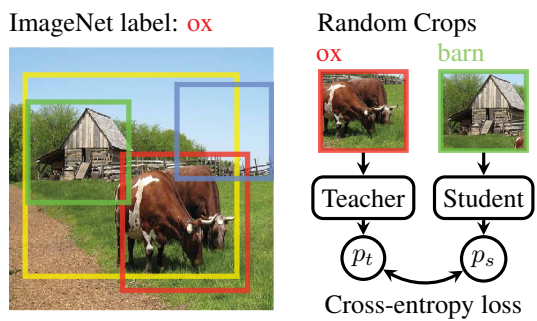
\begin{wrapfigure}{r}{0.51\textwidth}
\vspace{-0.3cm}
  \begin{center}
    \begin{tikzpicture}[
module/.style={draw, very thick, rounded corners, minimum
width=8ex, minimum height=3.5ex},
embmodule/.style={module, fill=red!20},
mhamodule/.style={module, fill=orange!20},
lnmodule/.style={module, fill=yellow!20},
ffnmodule/.style={module, fill=cyan!20},
arrow/.style={-stealth, very thick, rounded corners},
line/.style={-, thick, rounded corners},
]

\newcommand{\marginposemb}{0.3}
\newcommand{\marginadd}{0.1}
\newcommand{\marginffwd}{0.7}
\newcommand{\marginresleft}{0.3}
\newcommand{\ymarginpatchesvertical}{0.25}
\newcommand{\spwidth}{0.2}
\newcommand{\spheight}{0.2}
\newcommand{\patchwidth}{0.4}
\newcommand{\patchheight}{0.4}
\newcommand{\imagewidth}{1.8}
\newcommand{\imageheight}{1.6}
\newcommand{\xmargin}{0.8}
\newcommand{\ymargin}{-1.8}
\newcommand{\xpatches}{-1.5*\patchwidth}
\newcommand{\xmask}{\xpatches+1.7}

\node (label) at (1.35,3.8) {ImageNet label: {\color{Red} ox}};
\path[fill stretch image=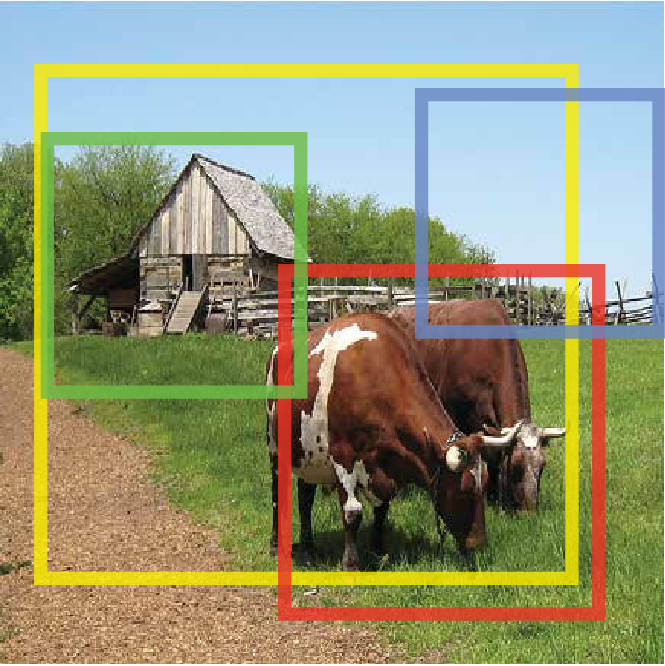] (0,0) rectangle (3.5,3.5);

\node[anchor=west,right=4.1 of label.west, minimum width=0cm,] (rc) {Random Crops};
\node (ox) at (4.25, 3.4) {{\color{Red} ox}};
\node (barn) at (6.0, 3.44) {{\color{YellowGreen} barn}};

\path[fill stretch image=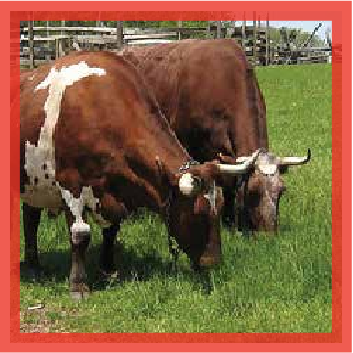] (4.1,2.1) rectangle (5.3,3.2);
\path[fill stretch image=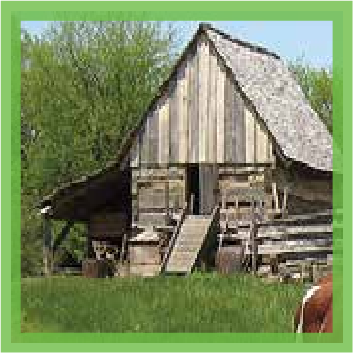] (5.7,2.1) rectangle (6.9,3.2);

\node[module, align=center] (teacher) at (4.7,1.5) {Teacher};
\node[module, align=center] (student) at (6.3,1.5) {Student};
\node[below=0.2 of teacher, draw, very thick, circle, inner sep=2pt, minimum size=1em] (pt) {$p_t$};
\node[below=0.2 of student, draw, very thick, circle, inner sep=2pt, minimum size=1em] (ps) {$p_s$};

\draw[very thick, stealth-stealth] (pt) to [out=330,in=210] (ps);
\node (cross) at (5.5, 0.05) [] {Cross-entropy loss};

\draw[arrow] (4.7,2.1) -- (teacher);
\draw[arrow] (6.3,2.1) -- (student);
\draw[arrow] (teacher) -- (pt);
\draw[arrow] (student) -- (ps);

\end{tikzpicture}
  \end{center}
  \vspace{-0.25cm}
  \caption{\textbf{(Left)} Multi-object example from ImageNet. Taken from \citep{yun2021re}.~\textbf{(Right)} Teacher and student see crops of distinct objects. 
  }
  \label{fig:IN-multi-obj}
  \vspace{-0.3cm}
\end{wrapfigure}
    At the image-level, self-distillation is implemented via a single-label classification objective on a global \texttt{[CLS]} embedding.
    While effective for object-centric datasets such as ImageNet \citep{deng2009imagenet}, 
    it implicitly assumes that
    \textit{each image contains a single object and the single-label objective can distill the most important content at the image level}.
    In practice, the image-level distillation loss funnels all information in the image into a single vector, entangling the semantics of co-occurring objects and background.
    This assumption mismatches the true data distribution for multi-object images. \looseness-1
    
    We highlight two key issues arising from the single-object assumption. First, even within ImageNet, recent works show that a significant fraction of images contain multiple objects as exemplified in \cref{fig:IN-multi-obj} (left) \citep{stock2018convnets,recht2019imagenet,tsipras2020imagenet,shankar2020evaluating,beyer2020we,yun2021re}. Indeed, roughly 20\% of images in ImageNet naturally require more than one label, reflected in improved multi-label validation set annotations \citep{tsipras2020imagenet,beyer2020we}, and improved multi-label training set annotations for supervised training \citep{yun2021re}.
    Yet, existing self-supervised pretraining approaches do not explicitly address such multi-object scenarios, e.g., random crops might contain distinct objects as in \cref{fig:IN-multi-obj} (right).
    Second, such image-level distillation does not directly scale to more complex, scene-centric datasets containing many interacting objects, where a single global representation overlooks valuable localized cues. \looseness-1

    This limitation is analogous to training language models exclusively on short, simple texts rather than long-form, context-rich corpora \citep{radford2018improving}. Just as language models see substantial gains when fed broader, more complex data, ViTs are expected to benefit from pretraining on scene-centric images or videos containing multiple objects. Realizing this goal, however, demands an approach that can accurately isolate and represent individual objects within complex scenes --a capability that has become increasingly feasible with modern segmentation models \citep{liu2024grounding,kirillov2023segment}. \looseness-1

    We address these challenges by introducing \textbf{O}bject-level Self-\textbf{Dis}tillation (ODIS), a pretraining method that refines self-distillation from the level of entire images to the granularity of individual objects (see \cref{fig:object-distillation}). By doing so, it transforms a noisy, complex scene-level task into simpler sub-tasks that focus on distinct entities. 
    ODIS explicitly guides the ViT toward more semantically meaningful object-specific content by \circled{1} \textbf{object-aware cropping} to ensure that the inputs to the student and teacher contain (different views of) the \emph{same} object, and \circled{2} \textbf{masked attention} to guide the optimization objective towards learning object-centric representations that are useful for downstream tasks such as classification. These observations can also be incorporated into contrastive learning \citep{chen2020simple}, masked image modeling \citep{he2022masked}, and multi-modal frameworks \citep{radford2021learning}. \looseness-1
    
    Empirically, ODIS significantly outperforms state-of-the-art image-level distillation methods on both image-level and patch-level benchmarks. Notably, a ViT-Large model pretrained with ODIS achieves $82.6\%$ $k$-NN accuracy on ImageNet1k when using masks at inference time, $+4.6\%$ improvement over iBOT \citep{zhou2021ibot} and $+0.6\%$ improvement over DINOv2 \citep{oquab2023dinov2}. Similarly, ODIS outperforms iBOT by a large margin even without segmentation masks at inference time, implying that our object-level distillation objective leads to better backbones. Beyond image-level classification gains, ODIS also boosts patch-level performance in an in-context scene understanding task \citep{balazevic2024towards}, highlighting the importance of moving beyond the single-object assumption and embracing multi-object pretraining in future vision foundation models.

\section{Related Work}
Below we summarize image- and object-level self-supervised learning methods. Please see \cref{appsec:relwork} for a review of the literature on object-centric learning and segmentation methods.

\textbf{Image-level self-supervision.} Inspired by the success of large-scale self-supervised pretraining in NLP, a large body of work has explored similar strategies for vision. Early approaches focused on pretext tasks such as masking and reconstructing patches \citep{he2022masked,bao2021beit}, potentially in feature space \citep{assran2023self}. These methods have demonstrated improved performance across diverse tasks when fine-tuned on specific downstream objectives. 

However, we focus on representations that are useful without additional fine-tuning, aligning more closely with discriminative image-level self-supervised methods  \citep{chen2020simple,grill2020bootstrap,caron2021emerging,zhou2021ibot,oquab2023dinov2}.
State-of-the-art methods typically employ a teacher-student framework  \citep{tarvainen2017mean} combined with image-level self-distillation \citep{grill2020bootstrap,caron2021emerging}, removing the necessity for negative examples \citep{chen2020simple}.
\citet{zhou2021ibot} combines the image-level self-distillation in \citep{caron2021emerging} with a patch-level loss inspired by masked language modeling \citep{devlin2019bert}.
Building on this, 
\citet{oquab2023dinov2} introduce algorithmic advances for stable large-scale training, and scale ViT pretraining to a 142M-image dataset and a 1B-parameter network. 
This led to state-of-the-art results in diverse vision tasks.
Our work also builds on iBOT \citep{zhou2021ibot}, however we focus on enhancing the learning objective to a finer level of granularity instead of scaling the pretraining.

\textbf{Object-level self-supervision.} A parallel line of research has investigated finer levels of granularity in self-supervised objectives, ranging from pixel level distillation \citep{o2020unsupervised} to patch-level \citep{wang2021dense} or full object-level \citep{henaff2021efficient,henaff2022object,xie2021unsupervised,stegmuller2023croc,wen2022self}.
These works primarily target dense downstream tasks such as object detection and semantic segmentation, and are often evaluated with either full fine-tuning \citep{henaff2021efficient,henaff2022object,wen2022self} or with a linear prediction head \citep{xie2021unsupervised,stegmuller2023croc}.\looseness-1

Closest to our approach are \citep{henaff2021efficient,henaff2022object}. Of particular interest, \citet{henaff2021efficient} formulates an object-level contrastive loss by leveraging object segmentation masks.
However, they employ average (linear) pooling over dense features to form object representations, limiting the expressivity of learned embeddings.
In contrast, our masked attention mechanism uses object segmentation masks at each transformer layer, yielding highly nonlinear object-level representations.
More importantly, while prior works emphasize fine-tuning for object detection and segmentation, our goal is to learn general-purpose object-level representations useful for downstream tasks out of the box.\looseness-1

\section{Preliminaries}
\label{sec:pre}

In this section, we briefly review the self-supervised pretraining algorithms of DINO \citep{caron2021emerging} and iBOT \citep{zhou2021ibot}, as our method builds on them.

\textbf{Input.} An input image $x \in \mathbb{R}^{C \times H_{\text{img}} \times W_{\text{img}}}$ is transformed via standard augmentations such as random cropping followed by a resize in order to obtain two random global views: $x^{(1)}, x^{(2)} \in \mathbb{R}^{C \times H_{\text{resize}} \times W_{\text{resize}}}$\footnote{For simplicity, we ignore local crops for now.}.
Two views $x^{(1)}$ and $x^{(2)}$ are divided into $H \times W$ patches and linearly projected to a $D$ dimensional embedding space: $\tilde{x}^{(1)}, \tilde{x}^{(2)} \in \mathbb{R}^{(HW) \times D}$. State-of-the-art pretraining approaches \citep{caron2021emerging,zhou2021ibot,oquab2023dinov2} typically concatenate the \texttt{[CLS]} $\in \mathbb{R}^{1 \times D}$ token which summarizes the image-level visual information: $[\texttt{[CLS]}, \tilde{x}] \in \mathbb{R}^{(1+HW) \times D}$.

\textbf{Network architecture.} The algorithm is implemented using a pair of student and teacher networks: $g_s = h_s \circ b_s$ and $g_t = h_t \circ b_t$, with ViT backbones $b_s, b_t$ and the MLP prediction heads $h_s, h_t$. The output activation of the MLP prediction heads $h_s, h_t$ are softmax with temperatures $t_s > t_t$.

\begin{figure}
    \centering
    \begin{subfigure}[b]{1.0\textwidth}
    \begin{tikzpicture}[scale=1.0]
        \newcommand{\ymarginpatchesvertical}{0.25}
        \newcommand{\spwidth}{0.2}
        \newcommand{\spheight}{0.2}
        \newcommand{\patchwidth}{0.4}
        \newcommand{\patchheight}{0.4}
        \newcommand{\imagewidth}{2.0}
        \newcommand{\imageheight}{1.6}
        \path[fill stretch image=figures/in-ex-crops.png] (-\imagewidth/2,-\imageheight/2) rectangle (\imagewidth/2,\imageheight/2);
        \node (rect) at (0,0) [draw,very thick,minimum width=\imagewidth cm,minimum height=\imageheight cm] {};
        \newcommand{\xmargin}{0.8}
        \newcommand{\ymargin}{0.6}
        \newcommand{\xpatches}{\imagewidth/2 + \xmargin+0.4}
        
        \path[fill stretch image=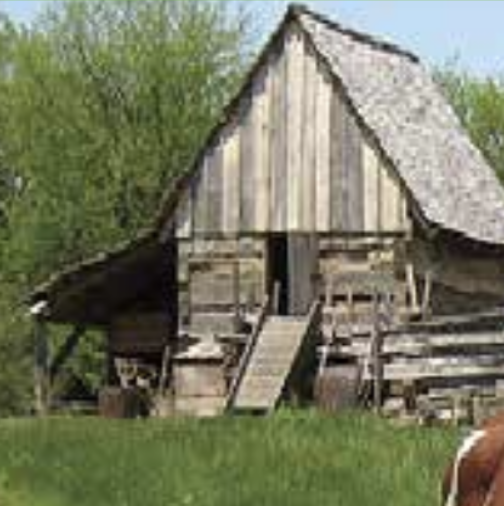] (\xpatches,\ymargin) rectangle (\xpatches+3*\patchwidth,\ymargin+3*\patchheight);
        \foreach \x in {0,1,2}{
            \foreach \y in {0,1,2}{
            \fill[fill=black!5, thick,opacity=0.35] (\xpatches+\x*\patchwidth, \ymargin+\y*\patchheight) rectangle (\xpatches+\x*\patchwidth+\patchwidth, \ymargin+\y*\patchheight+\patchheight);
            \draw[color=black, line width=1.2pt, opacity=1.0] (\xpatches+\x*\patchwidth, \ymargin+\y*\patchheight) rectangle (\xpatches+\x*\patchwidth+\patchwidth, \ymargin+\y*\patchheight+\patchheight);
            }
        }

        \node at (0,\imageheight/2+0.3) [] {Image $x \sim \mathcal{D}$};
        \filldraw[color=ForestGreen, fill=YellowGreen!30, very thick] (\xpatches+2*\patchwidth, \ymargin+3.3*\patchheight) rectangle (\xpatches+3*\patchwidth, \ymargin+4.3*\patchheight);
        \node at (\xpatches+1*\patchwidth-0.2, \ymargin+3.7*\patchheight) [] {View 1:};
        
        \filldraw[color=Red!90!black, fill=red!5, very thick] (\xpatches+2*\patchwidth, -\ymargin-3.3*\patchheight) rectangle (\xpatches+3*\patchwidth, -\ymargin-4.3*\patchheight);
        \node at (\xpatches+1*\patchwidth-0.2, -\ymargin-3.9*\patchheight) [] {View 2:};

        \path[fill stretch image=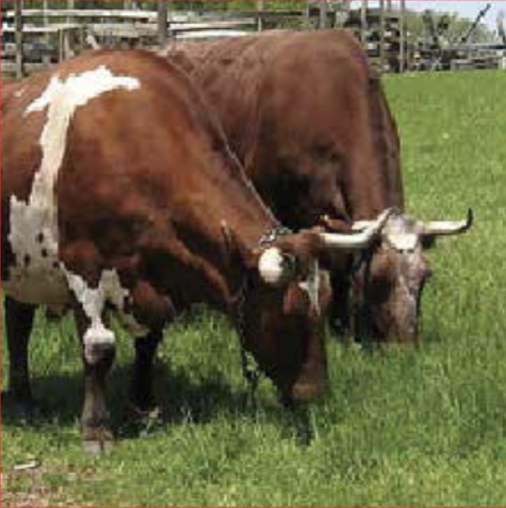] (\xpatches,-\ymargin) rectangle (\xpatches+3*\patchwidth,-\ymargin-3*\patchheight);
        \foreach \x in {0,1,2}{
            \foreach \y in {0,1,2}{
            \fill[fill=black!5, opacity=0.35] 
            (\xpatches+\x*\patchwidth, -\ymargin-\y*\patchheight) rectangle (\xpatches+\x*\patchwidth+\patchwidth, -\ymargin-\y*\patchheight-\patchheight);
            
            \draw[color=black, line width=1.2pt, opacity=1.0] 
            (\xpatches+\x*\patchwidth, -\ymargin-\y*\patchheight) rectangle (\xpatches+\x*\patchwidth+\patchwidth, -\ymargin-\y*\patchheight-\patchheight);
            }
        }
        
        \newcommand{\xpatchesend}{\xpatches+3*\patchwidth}  
        \newcommand{\ypatchesmidup}{\ymargin+\patchheight+\patchheight/2} 
        \newcommand{\ypatchesmiddown}{-\ymargin-\patchheight-\patchheight/2} 

        \draw[very thick, ->] (\imagewidth/2,0) -- (\xpatches,\ypatchesmidup);
        \draw[very thick, ->] (\imagewidth/2,0) -- (\xpatches,\ypatchesmiddown);

        \newcommand{\xpatchesvertical}{\xpatchesend+\xmargin+0.4}  
        \node at (\imagewidth/2,0) [anchor=west, rotate=0, xshift=0.18cm, yshift=0.18cm] {\small aug.\&};
        \node at (\imagewidth/2,0) [anchor=west, rotate=0, xshift=0.18cm, yshift=-0.12cm] {\small patchify};
        \node[anchor=south west] at (\xpatchesend+0.05,1.2) {\small flatten};
        \node[anchor=north west] at (\xpatchesend+0.05,-1.2) {\small flatten};
        
        \draw[very thick, ->] (\xpatchesend,1.2) -- (\xpatchesvertical,\ypatchesmidup);
        \draw[very thick, ->] (\xpatchesend,-1.2) -- (\xpatchesvertical,\ypatchesmiddown);
        
        \foreach \y in {0,1,...,8}{
            \filldraw[color=black, fill=black!10, very thick] (\xpatchesvertical, \ymarginpatchesvertical+\y*\spwidth) rectangle (\xpatchesvertical+\spheight, \ymarginpatchesvertical+\y*\spwidth+\spwidth);
        }
        \foreach \y in {0,1,...,8}{
            \filldraw[color=black, fill=black!10, very thick] (\xpatchesvertical, -\ymarginpatchesvertical-\y*\spwidth) rectangle (\xpatchesvertical+\spheight, -\ymarginpatchesvertical-\y*\spwidth-\spwidth);
        }

        \newcommand{\xcls}{\xpatchesvertical+\spwidth+0.35} 
        \newcommand{\yclsup}{4*\spwidth + \ymarginpatchesvertical}
        \newcommand{\yclsdown}{-4*\spwidth - \ymarginpatchesvertical}
        
        \filldraw[color=black, fill=black!10, very thick] (\xcls, \yclsup) rectangle (\xcls+\spheight, \yclsup+\spwidth);
        \node at (\xcls+0.4, \yclsup+\spwidth+0.3) [] {\texttt{[CLS]}};
        \filldraw[color=black, fill=black!10, very thick] (\xcls, \yclsdown) rectangle (\xcls+\spheight, \yclsdown-\spwidth);
        \node at (\xcls+0.4, \yclsdown-\spwidth-0.3) [] {\texttt{[CLS]}};
        
        \draw[very thick, -, color=black] (\xpatchesvertical+\spwidth,\ymarginpatchesvertical) -- (\xcls, \yclsup);
        \draw[very thick, -, color=black] (\xpatchesvertical+\spwidth,\ymarginpatchesvertical+\spheight*9) -- (\xcls, \yclsup+\spheight);

        \draw[very thick, -, color=black] (\xpatchesvertical+\spwidth,-\ymarginpatchesvertical) -- (\xcls, \yclsdown);
        \draw[very thick, -, color=black] (\xpatchesvertical+\spwidth,-\ymarginpatchesvertical-\spheight*9) -- (\xcls, \yclsdown-\spheight);
        

        \newcommand{\bwidth}{1.2}
        \newcommand{\bheight}{1.8}
        \newcommand{\xmarginnw}{\xcls+\spwidth+\spwidth+\xmargin+0.5*\bwidth-0.1}  
        \node (rect) at (\xmarginnw,\yclsup+\spheight/2) [draw,very thick,rounded corners,minimum width=\bwidth cm,minimum height=\bheight cm] {$b_t$};
        \node (rect) at (\xmarginnw,\yclsdown-\spheight/2) [draw,very thick,rounded corners,minimum width=\bwidth cm,minimum height=\bheight cm] {$b_s$};

        \newcommand{\hwidth}{1.2}
        \newcommand{\hheight}{0.8}
        \newcommand{\xmarginhead}{\xmarginnw+0.5*\bwidth+\xmargin+0.5*\hwidth-0.1}
        \node (rect) at (\xmarginhead,\yclsup+\spheight/2) [draw,very thick,rounded corners,minimum width=\hwidth cm,minimum height=\hheight cm] {$h_t$};
        \node (rect) at (\xmarginhead,\yclsdown-\spheight/2) [draw,very thick,rounded corners,minimum width=\hwidth cm,minimum height=\hheight cm] {$h_s$};
        
        \newcommand{\xmclsoutput}{\xmarginhead+\hwidth/2+\xmargin}
        \filldraw[color=black, fill=black!20, very thick] (\xmclsoutput, \yclsup) rectangle (\xmclsoutput+\spheight, \yclsup+\spwidth);
        \node at (\xmclsoutput+0.4, \yclsup+\spheight+0.35) [] {$p^{\texttt{[CLS]}}_t$};
        \filldraw[color=black, fill=black!20, very thick] (\xmclsoutput, \yclsdown) rectangle (\xmclsoutput+\spheight, \yclsdown-\spwidth);
        \node at (\xmclsoutput+0.4, \yclsdown-\spwidth-0.35) [] {$p^{\texttt{[CLS]}}_s$};

        \draw[very thick, ->] (\xcls+\spwidth,\yclsup+\spwidth/2) -- (\xmarginnw-0.5*\bwidth,\yclsup+\spwidth/2); 
        \draw[very thick, ->] (\xcls+\spwidth,\yclsdown-\spwidth/2) -- (\xmarginnw-0.5*\bwidth,\yclsdown-\spwidth/2);
        \draw[very thick, ->] (\xmarginnw+0.5*\bwidth,\yclsup+\spwidth/2) -- (\xmarginhead-0.5*\hwidth,\yclsup+\spwidth/2); 
        \draw[very thick, ->] (\xmarginnw+0.5*\bwidth,\yclsdown-\spwidth/2) -- (\xmarginhead-0.5*\hwidth,\yclsdown-\spwidth/2);
        \draw[very thick, ->] (\xmarginhead+0.5*\hwidth,\yclsup+\spwidth/2) -- (\xmclsoutput,\yclsup+\spwidth/2); 
        \draw[very thick, ->] (\xmarginhead+0.5*\hwidth,\yclsdown-\spwidth/2) -- (\xmclsoutput,\yclsdown-\spwidth/2);

        \newcommand{\xprobs}{\xmclsoutput+\spheight+\xmargin-0.1}
        \newcommand{\xprobsmargin}{0.15}
        \newcommand{\yprobsunitheight}{\bheight/7.5}
        \newcommand{\yprobs}{\yclsup+\spwidth/2-\bheight/2}
        \newcommand{\yprobsmax}{\yclsup+\spwidth/2+\bheight/2}
        \newcommand{\yprobsdown}{\yclsdown-\spwidth/2-\bheight/2}
        \newcommand{\yprobsdownmax}{\yclsdown-\spwidth/2+\bheight/2}
        
        \draw[ultra thick, color=black, -] (\xprobs, \yprobs) --  (\xprobs, \yprobsmax);
        \draw[ultra thick, color=black, -] (\xprobs, \yprobsdown) --  (\xprobs, \yprobsdownmax);

        \filldraw[color=black, fill=black!20, very thick] (\xprobs+\xprobsmargin, \yprobs+\yprobsunitheight) rectangle (\xprobs+\xprobsmargin+0.3, \yprobs+\yprobsunitheight+\yprobsunitheight);
        \filldraw[color=black, fill=black!20, very thick] (\xprobs+\xprobsmargin, \yprobs+2.5*\yprobsunitheight) rectangle (\xprobs+\xprobsmargin+0.1, \yprobs+2.5*\yprobsunitheight+\yprobsunitheight);
        \filldraw[color=black, fill=black!20, very thick] (\xprobs+\xprobsmargin, \yprobs+4*\yprobsunitheight) rectangle (\xprobs+\xprobsmargin+0.2, \yprobs+4*\yprobsunitheight+\yprobsunitheight);
        \filldraw[color=black, fill=black!20, very thick] (\xprobs+\xprobsmargin, \yprobs+5.5*\yprobsunitheight) rectangle (\xprobs+\xprobsmargin+0.5, \yprobs+5.5*\yprobsunitheight+\yprobsunitheight);

        \filldraw[color=black, fill=black!20, very thick] (\xprobs+\xprobsmargin, \yprobsdown+\yprobsunitheight) rectangle (\xprobs+\xprobsmargin+0.15, \yprobsdown+\yprobsunitheight+\yprobsunitheight);
        \filldraw[color=black, fill=black!20, very thick] (\xprobs+\xprobsmargin, \yprobsdown+2.5*\yprobsunitheight) rectangle (\xprobs+\xprobsmargin+0.05, \yprobsdown+2.5*\yprobsunitheight+\yprobsunitheight);
        \filldraw[color=black, fill=black!20, very thick] (\xprobs+\xprobsmargin, \yprobsdown+4*\yprobsunitheight) rectangle (\xprobs+\xprobsmargin+0.1, \yprobsdown+4*\yprobsunitheight+\yprobsunitheight);
        \filldraw[color=black, fill=black!20, very thick] (\xprobs+\xprobsmargin, \yprobsdown+5.5*\yprobsunitheight) rectangle (\xprobs+\xprobsmargin+0.7, \yprobsdown+5.5*\yprobsunitheight+\yprobsunitheight);
        
        \draw[very thick, stealth-stealth] (\xprobs+\xmargin+\spheight-0.3, \yclsup+\spwidth/2) to [out=330,in=30,] (\xprobs+\xmargin+\spheight-0.3, \yclsdown-\spwidth/2);
        \node at (\xprobs+\xmargin+\spheight+1.2-0.35, 0.0) [] {$\mathcal{L}_{\texttt{[CLS]}}$};

        \draw[decorate,decoration={brace,amplitude=5pt,mirror,raise=2pt},line width=1.5pt]
   (-\imagewidth/2, -\ymargin-4.8*\patchheight) --(\xpatchesend, -\ymargin-4.8*\patchheight) node[midway,yshift=-1.5em]{Random Cropping};\draw[decorate,decoration={brace,amplitude=5pt,mirror,raise=2pt},line width=1.5pt]
   (\xpatchesvertical, -\ymargin-4.8*\patchheight) --(\xmarginnw+\bwidth/2, -\ymargin-4.8*\patchheight) node[midway,yshift=-1.5em]{Full Attention};
        
        \end{tikzpicture}
    \vspace*{-0.3cm}
   \caption{Image-level self-distillation via \texttt{[CLS]} token with Random Cropping and Full Attention.}
   \vspace*{0.2cm}
   \label{fig:odis1} 
\end{subfigure}
\begin{subfigure}[b]{1.0\textwidth}
    \begin{tikzpicture}
        \newcommand{\ymarginpatchesvertical}{0.25}
        \newcommand{\spwidth}{0.2}
        \newcommand{\spheight}{0.2}
        \newcommand{\patchwidth}{0.4}
        \newcommand{\patchheight}{0.4}
        \newcommand{\imagewidth}{2.0}
        \newcommand{\imageheight}{1.6}

        \newcommand{\yimagemargin}{1.2}
        \path[fill stretch image=figures/in-ex-crops.png] (-\imagewidth/2,\yimagemargin-\imageheight/2) rectangle (\imagewidth/2,\yimagemargin+\imageheight/2);
        \node (rect) at (0,\yimagemargin) [draw,very thick,minimum width=\imagewidth cm,minimum height=\imageheight cm] {};

        \path[fill stretch image=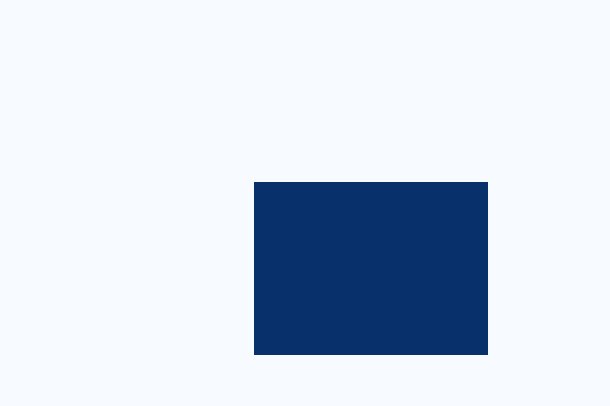] (-\imagewidth/2,-\yimagemargin-\imageheight/2) rectangle (\imagewidth/2,-\yimagemargin+\imageheight/2);
        \node (rect) at (0,-\yimagemargin) [draw,very thick,minimum width=\imagewidth cm,minimum height=\imageheight cm] {};
        
        
        \newcommand{\xmargin}{0.8}
        \newcommand{\ymargin}{0.6}
        \newcommand{\xpatches}{\imagewidth/2 + \xmargin+0.4}

        \node at (0,\yimagemargin+\imageheight/2+0.2) [] {Image $x \sim \mathcal{D}$};
        \node at (0,-\yimagemargin+\imageheight/2+0.2) [] {Object mask $y$};
        \draw[decorate,decoration={brace,amplitude=5pt,mirror,raise=2pt},line width=1.5pt]
   (\imagewidth/2+0.05, -\yimagemargin-\imageheight/2) --(\imagewidth/2+0.05,\yimagemargin+\imageheight/2);
        
        \filldraw[color=yellow!80!black, fill=yellow!20, very thick] (\xpatches+2*\patchwidth, \ymargin+4.6*\patchheight) rectangle (\xpatches+3*\patchwidth, \ymargin+5.6*\patchheight);
        \node at (\xpatches+1*\patchwidth-0.2, \ymargin+5*\patchheight) [] {View 1:};
        \filldraw[color=bluecorner, fill=bluefill, rounded corners, very thick] (\xpatches+2*\patchwidth, \ymargin+3.3*\patchheight) rectangle (\xpatches+3*\patchwidth, \ymargin+4.3*\patchheight);
        \node at (\xpatches+1*\patchwidth-0.1, \ymargin+3.8*\patchheight) [] {Mask:};
        
        \filldraw[color=bluecorner, fill=bluefill, rounded corners, very thick] (\xpatches+2*\patchwidth, -\ymargin-4.6*\patchheight) rectangle (\xpatches+3*\patchwidth, -\ymargin-5.6*\patchheight);
        \node at (\xpatches+1*\patchwidth-0.1, -\ymargin-5.2*\patchheight) [] {Mask:};
        \filldraw[color=Red!90!black, fill=red!5, very thick] (\xpatches+2*\patchwidth, -\ymargin-3.3*\patchheight) rectangle (\xpatches+3*\patchwidth, -\ymargin-4.3*\patchheight);
        \node at (\xpatches+1*\patchwidth-0.2, -\ymargin-3.9*\patchheight) [] {View 2:};
        
        \path[fill stretch image=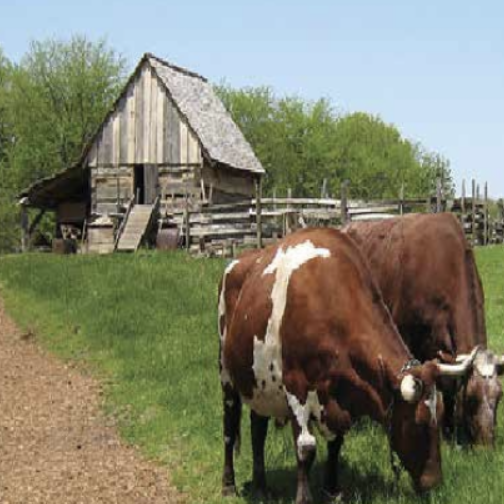] (\xpatches,\ymargin) rectangle (\xpatches+3*\patchwidth,\ymargin+3*\patchheight);
        
        \foreach \x in {0,1,2}{
            \foreach \y in {0,1,2}{
            \fill[fill=black!5, thick, opacity=0.3] (\xpatches+\x*\patchwidth, \ymargin+\y*\patchheight) rectangle (\xpatches+\x*\patchwidth+\patchwidth, \ymargin+\y*\patchheight+\patchheight);
            \draw[color=black, opacity=1.0, line width=1pt] (\xpatches+\x*\patchwidth, \ymargin+\y*\patchheight) rectangle (\xpatches+\x*\patchwidth+\patchwidth, \ymargin+\y*\patchheight+\patchheight);
            }
        }
        \draw[color=bluecorner, opacity=1.0, rounded corners, line width=3pt] (\xpatches+1*\patchwidth-0.05, \ymargin-0.05) rectangle (\xpatches+3*\patchwidth+0.05, \ymargin+1*\patchheight+\patchheight+0.05);
        \draw[color=bluecorner, fill=Cyan, opacity=0.2, rounded corners, line width=3pt] (\xpatches+1*\patchwidth-0.05, \ymargin-0.05) rectangle (\xpatches+3*\patchwidth+0.05, \ymargin+1*\patchheight+\patchheight+0.05);
        
        \path[fill stretch image=figures/in-ex-crop1.png] (\xpatches,-\ymargin) rectangle (\xpatches+3*\patchwidth,-\ymargin-3*\patchheight);
        \foreach \x in {0,1,2}{
            \foreach \y in {0,1,2}{
            \fill[fill=black!5, thick, opacity=0.3] 
            (\xpatches+\x*\patchwidth, -\ymargin-\y*\patchheight) rectangle (\xpatches+\x*\patchwidth+\patchwidth, -\ymargin-\y*\patchheight-\patchheight);
            
            \draw[color=black, opacity=1.0, line width=1pt] 
            (\xpatches+\x*\patchwidth, -\ymargin-\y*\patchheight) rectangle (\xpatches+\x*\patchwidth+\patchwidth, -\ymargin-\y*\patchheight-\patchheight);
            }
        }
        \draw[color=bluecorner, rounded corners, very thick, line width=3pt, opacity=1.0] (\xpatches-0.05, -\ymargin+0.05) rectangle (\xpatches+3*\patchwidth+0.05, -\ymargin-2*\patchheight-0.05);
        \draw[color=bluecorner, rounded corners, fill=Cyan, very thick, line width=3pt, opacity=0.2] (\xpatches-0.05, -\ymargin+0.05) rectangle (\xpatches+3*\patchwidth+0.05, -\ymargin-2*\patchheight-0.05);
        
        \newcommand{\xpatchesend}{\xpatches+3*\patchwidth}  
        
        \newcommand{\ypatchesmidup}{\ymargin+\patchheight+\patchheight/2} 
        \newcommand{\ypatchesmiddown}{-\ymargin-\patchheight-\patchheight/2} 
        \draw[very thick, ->] (\imagewidth/2+0.4,0) -- (\xpatches,\ypatchesmidup);
        \draw[very thick, ->] (\imagewidth/2+0.4,0) -- (\xpatches-0.05,\ypatchesmiddown);

        \newcommand{\xpatchesvertical}{\xpatchesend+\xmargin+0.4}  
        \node at (\imagewidth/2,0) [anchor=west, rotate=0, xshift=0.47cm, yshift=0.17cm] {\small aug.\&};
        \node at (\imagewidth/2,0) [anchor=west, rotate=0, xshift=0.47cm, yshift=-0.13cm] {\small patchify};
        \node[anchor=south west] at (\xpatchesend+0.1,1.2) {\small flatten};
        \node[anchor=north west] at (\xpatchesend+0.1,-1.2) {\small flatten};
        
        \draw[very thick, ->] (\xpatchesend,1.2) -- (\xpatchesvertical,\ypatchesmidup);
        \draw[very thick, ->] (\xpatchesend,-1.2) -- (\xpatchesvertical,\ypatchesmiddown);
        
        \foreach \y in {8,7,6,5,2}{
            \filldraw[color=black, fill=black!10, very thick] (\xpatchesvertical, \ymarginpatchesvertical+\y*\spwidth) rectangle (\xpatchesvertical+\spheight, \ymarginpatchesvertical+\y*\spwidth+\spwidth);
        }
        \foreach \y in {4,3,1,0}{
            \filldraw[color=bluecorner, fill=bluefill, very thick] (\xpatchesvertical, \ymarginpatchesvertical+\y*\spwidth) rectangle (\xpatchesvertical+\spheight, \ymarginpatchesvertical+\y*\spwidth+\spwidth);
        }
        \foreach \y in {6,7,8}{
            \filldraw[color=black, fill=black!10, very thick] (\xpatchesvertical, -\ymarginpatchesvertical-\y*\spwidth) rectangle (\xpatchesvertical+\spheight, -\ymarginpatchesvertical-\y*\spwidth-\spwidth);
        }
        \foreach \y in {0,1,...,5}{
            \filldraw[color=bluecorner, fill=bluefill, very thick] (\xpatchesvertical, -\ymarginpatchesvertical-\y*\spwidth) rectangle (\xpatchesvertical+\spheight, -\ymarginpatchesvertical-\y*\spwidth-\spwidth);
        }

        \newcommand{\xcls}{\xpatchesvertical+\spwidth+0.35} 
        \newcommand{\yclsup}{4*\spwidth + \ymarginpatchesvertical}
        \newcommand{\yclsdown}{-4*\spwidth - \ymarginpatchesvertical}
        
        \filldraw[color=bluecorner, fill=bluefill, very thick] (\xcls, \yclsup) rectangle (\xcls+\spheight, \yclsup+\spwidth);
        \node at (\xcls+0.4, \yclsup+\spwidth+0.3) [] {\texttt{[OBJ]}};
        \filldraw[color=bluecorner, fill=bluefill, very thick] (\xcls, \yclsdown) rectangle (\xcls+\spheight, \yclsdown-\spwidth);
        \node at (\xcls+0.4, \yclsdown-\spwidth-0.3) [] {\texttt{[OBJ]}};
        
        \draw[very thick, -, color=bluecorner] (\xpatchesvertical+\spwidth,\ymarginpatchesvertical) -- (\xcls, \yclsup);
        \draw[very thick, -, color=bluecorner] (\xpatchesvertical+\spwidth,\ymarginpatchesvertical) -- (\xcls, \yclsup);
        \draw[very thick, -, color=bluecorner] (\xpatchesvertical+\spwidth,\ymarginpatchesvertical+\spheight*3) -- (\xcls, \yclsup);
        
        \draw[very thick, -, color=bluecorner] (\xpatchesvertical+\spwidth,\ymarginpatchesvertical+\spheight*2) -- (\xcls, \yclsup+\spheight);
        \draw[very thick, -, color=bluecorner] (\xpatchesvertical+\spwidth,\ymarginpatchesvertical+\spheight*5) -- (\xcls, \yclsup+\spheight);

        \draw[very thick, -, color=bluecorner] (\xpatchesvertical+\spwidth,-\ymarginpatchesvertical) -- (\xcls, \yclsdown);
        \draw[very thick, -, color=bluecorner] (\xpatchesvertical+\spwidth,-\ymarginpatchesvertical-\spheight*6) -- (\xcls, \yclsdown-\spheight);
        

        \newcommand{\bwidth}{1.2}
        \newcommand{\bheight}{1.8}
        \newcommand{\xmarginnw}{\xcls+\spwidth+\spwidth+\xmargin+0.5*\bwidth-0.1}  
        
        \node (rect) at (\xmarginnw,\yclsup+\spheight/2) [draw,very thick,rounded corners,minimum width=\bwidth cm,minimum height=\bheight cm] {$b_t$};
        \node (rect) at (\xmarginnw,\yclsdown-\spheight/2) [draw,very thick,rounded corners,minimum width=\bwidth cm,minimum height=\bheight cm] {$b_s$};

        \draw[decorate,decoration={brace,amplitude=5pt,mirror,raise=2pt},line width=1.5pt]
   (-\imagewidth/2, -\ymargin-5.8*\patchheight) --(\xpatchesend, -\ymargin-5.8*\patchheight) node[midway,yshift=-1.5em]{\circled{1} Object-aware Cropping};\draw[decorate,decoration={brace,amplitude=5pt,mirror,raise=2pt},line width=1.5pt]
   (\xpatchesvertical, -\ymargin-5.8*\patchheight) --(\xmarginnw+\bwidth/2, -\ymargin-5.8*\patchheight) node[midway,yshift=-1.5em]{\circled{2} Masked Attention};

        \newcommand{\hwidth}{1.2}
        \newcommand{\hheight}{0.8}
        \newcommand{\xmarginhead}{\xmarginnw+0.5*\bwidth+\xmargin+0.5*\hwidth-0.1}
        
        \node (rect) at (\xmarginhead,\yclsup+\spheight/2) [draw,very thick,rounded corners,minimum width=\hwidth cm,minimum height=\hheight cm] {$h_t$};
        \node (rect) at (\xmarginhead,\yclsdown-\spheight/2) [draw,very thick,rounded corners,minimum width=\hwidth cm,minimum height=\hheight cm] {$h_s$};
        
        \newcommand{\xmclsoutput}{\xmarginhead+\hwidth/2+\xmargin-0.1}
        \filldraw[color=bluecorner, fill=bluefill, very thick] (\xmclsoutput, \yclsup) rectangle (\xmclsoutput+\spheight, \yclsup+\spwidth);
        \node at (\xmclsoutput+0.4, \yclsup+\spheight+0.35) [] {$p^{\texttt{[OBJ]}}_t$};
        \filldraw[color=bluecorner, fill=bluefill, very thick] (\xmclsoutput, \yclsdown) rectangle (\xmclsoutput+\spheight, \yclsdown-\spwidth);
        \node at (\xmclsoutput+0.4, \yclsdown-\spwidth-0.35) [] {$p^{\texttt{[OBJ]}}_s$};

        \draw[very thick, ->] (\xcls+\spwidth,\yclsup+\spwidth/2) -- (\xmarginnw-0.5*\bwidth,\yclsup+\spwidth/2);
        \draw[very thick, ->] (\xcls+\spwidth,\yclsdown-\spwidth/2) -- (\xmarginnw-0.5*\bwidth,\yclsdown-\spwidth/2);
        \draw[very thick, ->] (\xmarginnw+0.5*\bwidth,\yclsup+\spwidth/2) -- (\xmarginhead-0.5*\hwidth,\yclsup+\spwidth/2); 
        \draw[very thick, ->] (\xmarginnw+0.5*\bwidth,\yclsdown-\spwidth/2) -- (\xmarginhead-0.5*\hwidth,\yclsdown-\spwidth/2);
        \draw[very thick, ->] (\xmarginhead+0.5*\hwidth,\yclsup+\spwidth/2) -- (\xmclsoutput,\yclsup+\spwidth/2); 
        \draw[very thick, ->] (\xmarginhead+0.5*\hwidth,\yclsdown-\spwidth/2) -- (\xmclsoutput,\yclsdown-\spwidth/2);

        \newcommand{\xprobs}{\xmclsoutput+\spheight+\xmargin-0.1}
        \newcommand{\xprobsmargin}{0.15}
        \newcommand{\yprobsunitheight}{\bheight/7.5}
        \newcommand{\yprobs}{\yclsup+\spwidth/2-\bheight/2}
        \newcommand{\yprobsmax}{\yclsup+\spwidth/2+\bheight/2}
        \newcommand{\yprobsdown}{\yclsdown-\spwidth/2-\bheight/2}
        \newcommand{\yprobsdownmax}{\yclsdown-\spwidth/2+\bheight/2}
        
        \draw[ultra thick, color=bluecorner, -] (\xprobs, \yprobs) --  (\xprobs, \yprobsmax);
        \draw[ultra thick, color=bluecorner, -] (\xprobs, \yprobsdown) --  (\xprobs, \yprobsdownmax);

        \filldraw[color=bluecorner, fill=bluefill, very thick] (\xprobs+\xprobsmargin, \yprobs+\yprobsunitheight) rectangle (\xprobs+\xprobsmargin+0.3, \yprobs+\yprobsunitheight+\yprobsunitheight);
        \filldraw[color=bluecorner, fill=bluefill, very thick] (\xprobs+\xprobsmargin, \yprobs+2.5*\yprobsunitheight) rectangle (\xprobs+\xprobsmargin+0.1, \yprobs+2.5*\yprobsunitheight+\yprobsunitheight);
        \filldraw[color=bluecorner, fill=bluefill, very thick] (\xprobs+\xprobsmargin, \yprobs+4*\yprobsunitheight) rectangle (\xprobs+\xprobsmargin+0.2, \yprobs+4*\yprobsunitheight+\yprobsunitheight);
        \filldraw[color=bluecorner, fill=bluefill, very thick] (\xprobs+\xprobsmargin, \yprobs+5.5*\yprobsunitheight) rectangle (\xprobs+\xprobsmargin+0.5, \yprobs+5.5*\yprobsunitheight+\yprobsunitheight);

        \filldraw[color=bluecorner, fill=bluefill, very thick] (\xprobs+\xprobsmargin, \yprobsdown+\yprobsunitheight) rectangle (\xprobs+\xprobsmargin+0.15, \yprobsdown+\yprobsunitheight+\yprobsunitheight);
        \filldraw[color=bluecorner, fill=bluefill, very thick] (\xprobs+\xprobsmargin, \yprobsdown+2.5*\yprobsunitheight) rectangle (\xprobs+\xprobsmargin+0.05, \yprobsdown+2.5*\yprobsunitheight+\yprobsunitheight);
        \filldraw[color=bluecorner, fill=bluefill, very thick] (\xprobs+\xprobsmargin, \yprobsdown+4*\yprobsunitheight) rectangle (\xprobs+\xprobsmargin+0.1, \yprobsdown+4*\yprobsunitheight+\yprobsunitheight);
        \filldraw[color=bluecorner, fill=bluefill, very thick] (\xprobs+\xprobsmargin, \yprobsdown+5.5*\yprobsunitheight) rectangle (\xprobs+\xprobsmargin+0.7, \yprobsdown+5.5*\yprobsunitheight+\yprobsunitheight);
        
        \draw[very thick, stealth-stealth] (\xprobs+\xmargin+\spheight-0.3, \yclsup+\spwidth/2) to [out=330,in=30,] (\xprobs+\xmargin+\spheight-0.3, \yclsdown-\spwidth/2);
        \node at (\xprobs+\xmargin+\spheight+1.2-0.35, 0.0) [] {$\mathcal{L}_{\texttt{[OBJ]}}$};
        
        \end{tikzpicture}
    \vspace*{-0.3cm}
   \caption{Object-level self-distillation via \texttt{[OBJ]} token with Object-aware Cropping and Masked Attention.}
   \label{fig:odis2} 
\end{subfigure}
    \caption{\textbf{Image-level vs.~Object-level distillation.} \textbf{(a)} Standard random cropping have no inherent mechanism to ensure that the student and teacher receive the same object as input. Hence, the distilled \texttt{[CLS]} tokens may summarize semantically different entities. \textbf{(b)} Our approach resolves this issue by \protect\circled{1} Object-aware Cropping that uses object masks. Further, \protect\circled{2} Masked Attention guides the \texttt{[OBJ]} token to pool information only from object tokens, leading to better representations.} 
    \label{fig:object-distillation}
\end{figure}
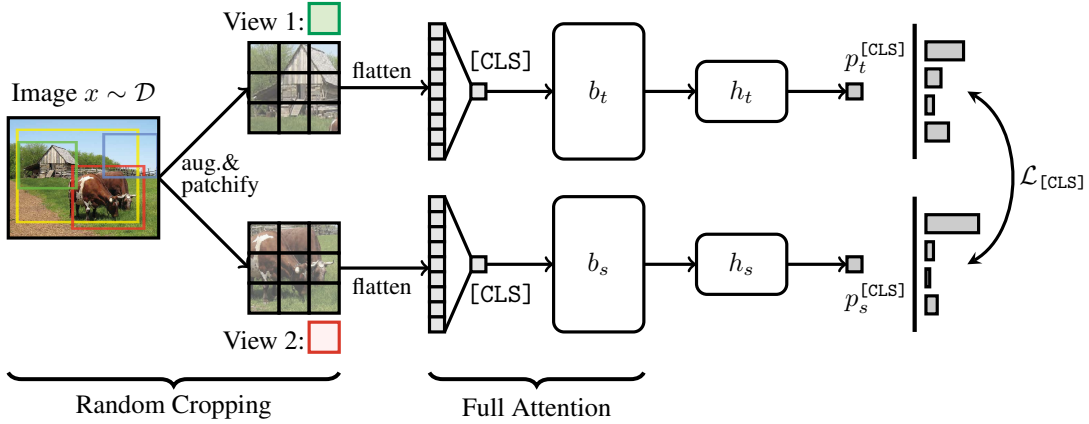
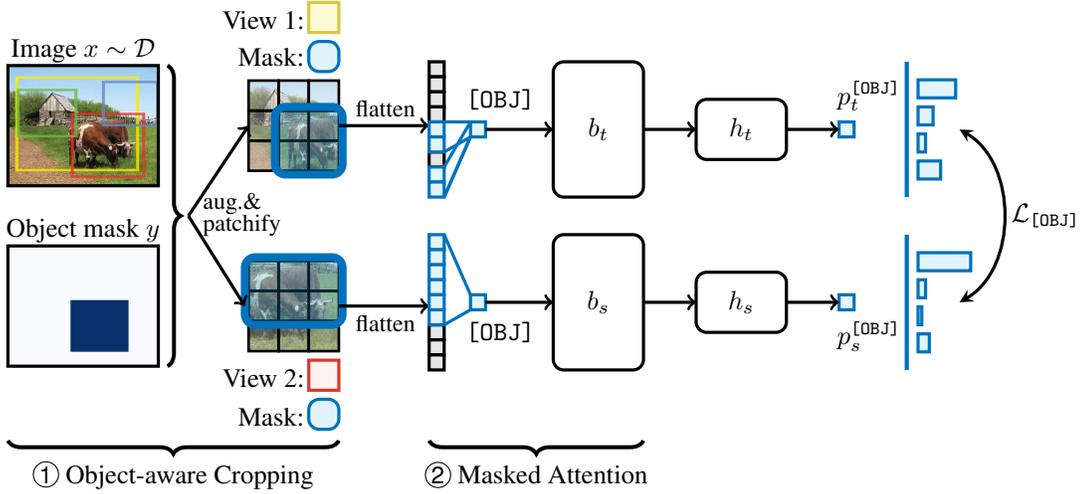

\textbf{Visual representations.} Visual representations are the outputs of the ViT backbones.
Although both the teacher and student process both global views in practice, for clarity we illustrate a simplified scenario where the teacher receives \colorboxtight{yellowfill}{\texttt{view 1}} and the student receives \colorboxtight{redfill}{\texttt{view 2}} (using the view color coding in \cref{fig:odis2}).
\newcommand{\uupper}{\colormathbox{yellowfill}{(1)}}
\newcommand{\vupper}{\colormathbox{redfill}{(2)}}
\begin{align}
    z^{\uupper}_{\texttt{[CLS]},t}, z^{\uupper}_{\texttt{patches},t} &= b_t ([\texttt{[CLS]}^{\colormathbox{yellowfill}{(1)}}, \tilde{x}^{\colormathbox{yellowfill}{(1)}}]), & \texttt{teacher - \colorboxtight{yellowfill}{view 1}} \\
    z^{\vupper}_{\texttt{[CLS]},s}, z^{\vupper}_{\texttt{patches},s} &= b_s ([\texttt{[CLS]}^{\vupper}, \tilde{x}^{\vupper}]), & \texttt{student - \colorboxtight{redfill}{view 2}}
\end{align}
with image-level representation $z_{\texttt{[CLS]}} \in \mathbb{R}^{1 \times D}$ and patch-level representations $z_{\texttt{patches}} \in \mathbb{R}^{HW \times D}$.

\textbf{Image-level objective (DINO Loss) \citep{caron2021emerging}.} MLP heads take the representations $z_{\texttt{[CLS]}}$ as input and produce probability vectors $p_s, p_t$, e.g., $p_{\texttt{[CLS]},s} = h_s(z_{\texttt{[CLS]}})$. We take \texttt{CrossEntropy} (\texttt{CE}) loss between probability vectors $p_s, p_t$ that correspond to distinct views $x^{\uupper}, x^{\vupper}$\footnote{Cross-entropy is defined as the dot product: $\texttt{CrossEntropy}(p_{\texttt{[CLS]},t}^{\uupper}, p_{\texttt{[CLS]},s}^{\vupper})= [p^{\uupper}_{\texttt{[CLS]},t}]^T [\log p_{\texttt{[CLS]},s}^{\vupper}]$.}:
\begin{align}
    p_{\texttt{[CLS]},t}^{\uupper} &= h_t(z^{\uupper}_{\texttt{[CLS]}}), & \texttt{teacher - \colorboxtight{yellowfill}{view 1} [CLS]} \\
    p_{\texttt{[CLS]},s}^{\vupper} &= h_s(z^{\vupper}_{\texttt{[CLS]}}), & \texttt{student - \colorboxtight{redfill}{view 2} [CLS]} \\
    \mathcal{L}_{\texttt{[CLS]}} &= \texttt{CrossEntropy}(p_{\texttt{[CLS]},t}^{\uupper}, p_{\texttt{[CLS]},s}^{\vupper}), & \texttt{DINO loss}
\end{align}
For clarity, we only provided the loss term for the simplified scenario above. The full loss is symmetric across views: $\mathcal{L}_{\texttt{[CLS]}} = \frac{1}{2}(\texttt{CE}(p_{\texttt{[CLS]},t}^{\uupper}, p_{\texttt{[CLS]},s}^{\vupper}) + \texttt{CE}(p_{\texttt{[CLS]},t}^{\vupper}, p_{\texttt{[CLS]},s}^{\uupper}))$.

\textbf{Patch-level objective and iBOT loss \citep{zhou2021ibot}.} iBOT creates an additional masked-image modeling task. For the student network input, it applies a random binary mask $m_1 \in \{0,1\}^{HW}$ to the input patch tokens $\tilde{x}^{(1)},\tilde{x}^{(2)} \in \mathbb{R}^{(HW) \times D}$. The masking replaces corresponding tokens by a general \texttt{[PATCH]} token, e.g., $\tilde{x}^{(1)}[m_1] := \texttt{[PATCH]}$\footnote{We use \texttt{torch} boolean mask notation in $\tilde{x}^{(1)}[m_1]$, selecting entries $i \in [HW]$ in $\tilde{x}^{(1)}[i]$ when $m_1[i] = 1$.}. The teacher receives unmasked patch tokens. Similar to image-level loss, MLP prediction heads produce probability vectors, e.g., $p_{\texttt{patches},s} = h_s(z_{\texttt{patches},s})$. In contrast to the cross-view formulation of the image-level loss, the patch-level loss is computed as follows for a single patch with patch index $i \in [HW]$ corresponding to the same views:
\begin{align}
    p_{\texttt{patches},t}^{\uupper} &= h_t(z^{\uupper}_{\texttt{patches}}), & \scalebox{0.9}{\texttt{teacher -
    \colorboxtight{yellowfill}{\small view 1} 
    unmask.}} \\
    p_{\texttt{patches},s}^{\uupper} &= h_s(z^{\uupper}_{\texttt{patches}}), & \scalebox{0.9}{\texttt{student -
    \colorboxtight{yellowfill}{\small view 1} 
    mask.}} \\
    \mathcal{L}_{\texttt{[PATCH]}} [i] &= m_1[i]~\texttt{CrossEntropy}(p_{\texttt{patches},t}^{\uupper}[i], p_{\texttt{patches},s}^{\uupper}[i]) & \scalebox{0.9}{\texttt{patch loss for } $i \in [HW]$}
\end{align}
which is summed over all masked patches: $\mathcal{L}_{\texttt{[PATCH]}} = - \frac{1}{\sum_j m(j)} \sum_{i \in [HW]} \mathcal{L}_{\texttt{[PATCH]}} [i]$. The iBOT loss sums up image- and patch-level losses: $\mathcal{L}_{\text{iBOT}} = \mathcal{L}_{\texttt{[CLS]}} + \mathcal{L}_{\texttt{[PATCH]}}$.

\textbf{Optimization.} The student network parameters $\theta_s$ are updated at every step via stochastic gradient descent. The gradients do not flow back to the teacher network, instead the teacher parameters $\theta_t$ are updated at every epoch as an exponential moving average (EMA) of the student parameters $\theta_s$: $\theta_t = \lambda \theta_t + (1-\lambda) \theta_s$ \citep{tarvainen2017mean}.

\section{Object-Level Self-Distillation}
\begin{wrapfigure}{r}{0.34\textwidth}
\vspace{-1.9cm}
    \centering
    \begin{tikzpicture}[
module/.style={draw, very thick, rounded corners, minimum
width=20ex},
embmodule/.style={module, fill=red!20},
mhamodule/.style={module, fill=orange!20},
lnmodule/.style={module, fill=yellow!20},
ffnmodule/.style={module, fill=cyan!20},
arrow/.style={-stealth, very thick, rounded corners},
line/.style={-, very thick, rounded corners},
]
\fontfamily{qcr}\selectfont

\newcommand{\marginposemb}{0.3}
\newcommand{\marginadd}{0.1}
\newcommand{\marginffwd}{0.7}
\newcommand{\marginresleft}{0.3}
\newcommand{\ymarginpatchesvertical}{0.25}
\newcommand{\spwidth}{0.2}
\newcommand{\spheight}{0.2}
\newcommand{\patchwidth}{0.4}
\newcommand{\patchheight}{0.4}
\newcommand{\imagewidth}{1.8}
\newcommand{\imageheight}{1.6}

\node (inputs) {Patches};
\newcommand{\xmargin}{0.8}
\newcommand{\ymargin}{-1.8}
\newcommand{\xpatches}{-1.5*\patchwidth}
\newcommand{\xmask}{\xpatches+1.7}

\path[fill stretch image=figures/in-ex-crop3.png] (\xpatches,\ymargin) rectangle (\xpatches+3*\patchwidth,\ymargin+3*\patchheight);
\foreach \x in {0,1,2}{
    \foreach \y in {0,1,2}{
    \fill[fill=black!5, thick,opacity=0.35] (\xpatches+\x*\patchwidth, \ymargin+\y*\patchheight) rectangle (\xpatches+\x*\patchwidth+\patchwidth, \ymargin+\y*\patchheight+\patchheight);
    \draw[color=black, line width=1.2pt, opacity=1.0] (\xpatches+\x*\patchwidth, \ymargin+\y*\patchheight) rectangle (\xpatches+\x*\patchwidth+\patchwidth, \ymargin+\y*\patchheight+\patchheight);
    }
}
\node[right=1.8 of inputs.west] (mask) {Object};
\node[below=0.35 of mask.north] (mask1) {Mask};
\path[fill stretch image=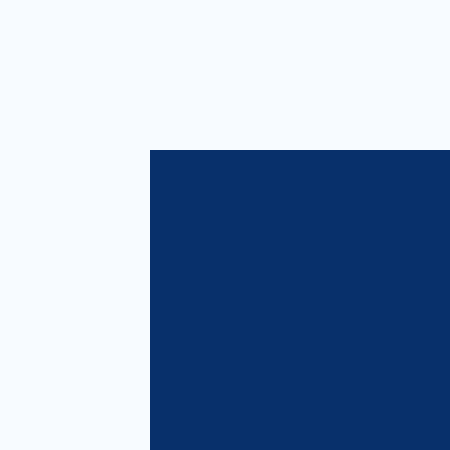] (\xmask,\ymargin) rectangle (\xmask+3*\patchwidth,\ymargin+3*\patchheight);
\foreach \x in {0,1,2}{
    \foreach \y in {0,1,2}{
    \fill[fill=black!5, thick,opacity=0.35] (\xmask+\x*\patchwidth, \ymargin+\y*\patchheight) rectangle (\xmask+\x*\patchwidth+\patchwidth, \ymargin+\y*\patchheight+\patchheight);
    \draw[color=black, line width=1.2pt, opacity=1.0] (\xmask+\x*\patchwidth, \ymargin+\y*\patchheight) rectangle (\xmask+\x*\patchwidth+\patchwidth, \ymargin+\y*\patchheight+\patchheight);
    }
}

\node[above=\marginposemb of inputs, embmodule, align=center, minimum
width=8ex] (inputemb) {Input\\Embed};
\node[above=\marginposemb of inputemb, draw, thick, circle, inner sep=0, minimum size=1em] (embplus)
{$\boldsymbol{+}$};
\node[above=1.75 of inputemb.north west, anchor=south west, mhamodule, align=center] (mha)
{Masked\\Multi-Head\\Attention};
\node[left=\marginposemb of embplus, draw, thick, circle, inner
sep=0pt,label={[align=left]left:Pos.\\Enc.}]
(pe) {\tikz \draw[scale=0.1,thick] plot[domain=0.0:6.28]
(\x,{sin(\x r)});};
\node[above=\marginadd of mha, lnmodule, align=center] (addnorm1)
{Add \& Norm};
\node[above=\marginffwd of addnorm1, ffnmodule, align=center] (ffn)
{Feed\\Forward};
\node[above=\marginadd of ffn, lnmodule, align=center] (addnorm2)
{Add \& Norm};
\node[above=of addnorm2] (outputs) {Outputs};

\coordinate (mharightanchor) at ($(mha.south)!0.595!(mha.south east)$);
\coordinate (mhaleftanchor) at ($(mha.south east)!0.77!(mha.south west)$);
\coordinate (mharesidual) at ($(mhaleftanchor)!0.5!(embplus.north)$);
\coordinate (ffnresidual) at ($(ffn.south)!0.5!(addnorm1.north)$);
\coordinate (mhafork) at ($(mhaleftanchor)!0.5!(mharesidual)$);
\coordinate[left=\marginresleft of addnorm1] (ln1residualleft);
\coordinate[left=\marginresleft of addnorm2] (ln2residualleft);

\draw[arrow] (inputs) -- (inputemb);
\draw[arrow] (inputemb) -- (embplus);
\draw[line] (pe) -- (embplus);
\draw[arrow] (embplus) -- (mhaleftanchor);
\draw[line] (mha) -- (addnorm1);
\draw[arrow] (addnorm1) -- (ffn);
\draw[line] (ffn) -- (addnorm2);
\draw[arrow] (addnorm2) -- (outputs);
\draw[arrow] (mask) -- (mharightanchor);

\node[fit={(mha)(addnorm2)(mharesidual)(ln1residualleft)}, draw, ultra thick, rounded corners, label=left:$\mathrm{N\times}$] (encoder) {};

\draw[arrow] (mharesidual)-|(ln1residualleft)--(addnorm1);
\draw[arrow] (ffnresidual)-|(ln2residualleft)--(addnorm2);
\draw[arrow] (mhafork)-|($(mha.south east)!0.90!(mha.south west)$);
\draw[arrow] (mhafork)-|($(mha.south east)!0.64!(mha.south west)$);

\end{tikzpicture}
    \caption{\textbf{Masked Attention with Object Segmentation Masks.}}
    \label{fig:masked-attention}
\vspace{-1.1cm}
\end{wrapfigure}
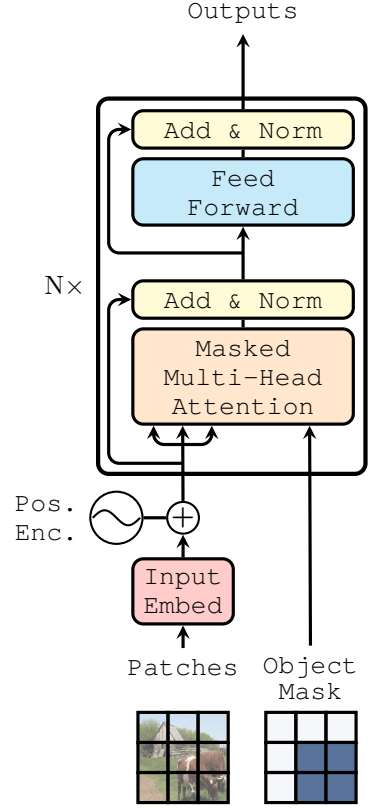
Next, we detail \textbf{O}bject-level Self-\textbf{Dis}tillation (ODIS), our proposed pretraining method that redefines self-distillation at the object level rather than the conventional image level. ODIS is built around two key components: \circled{1} object-aware cropping, which ensures that both student and teacher networks receive distinct views of the same object, and \circled{2} masked attention, which focuses the learning objective on objects, illustrated in \cref{fig:object-distillation,fig:masked-attention}. Together, these components guide the model toward learning richer, object-centric representations that transfer effectively to downstream tasks such as classification.

\textbf{\protect\circled{1} Object-aware cropping} 
In addition to an input image $x \in \mathbb{R}^{C \times H_{\text{img}} \times W_{\text{img}}}$, the model also receives a binary object segmentation map $y \in \{0,1\}^{H_{\text{img}} \times W_{\text{img}}}$, where $y_{ij} = 1$ if the object is present at pixel location $(i,j)$ (our method equivalently works with bounding boxes). While augmenting the input image $x$ to obtain two random views $x^{\uupper}, x^{\vupper} \in \mathbb{R}^{C \times H_{\text{resize}} \times W_{\text{resize}}}$, we apply the same spatial transformations to the object segmentation map $y$ to obtain two segmentation views aligned with the image views: $y^{\uupper}, y^{\vupper} \in \{0,1\}^{H_{\text{resize}} \times W_{\text{resize}}}$.
Similar to image views, the segmentation views $y^{\uupper}, y^{\vupper}$ are further divided into $H \times W$ patches, and transformed into binary object masks $\tilde{y}^{\uupper}, \tilde{y}^{\vupper} \in \{0,1\}^{HW}$ 
We ensure that the target object is present in both global views by randomly cropping up to 20 times and keeping the global views that contain the target object. 

Depending on the dataset, an image might contain multiple objects, and multiple object locations might be annotated as segmentation maps.
When an input segmentation map includes multiple distinct objects during training, we sample a single target object per forward pass.
To sample the target object, we consider two object sampling strategies:  at random or at random proportional to object areas (see ablations for details).
This way, the model targets a single object per forward pass while being able to see all objects in an image throughout training epochs.

\textbf{\protect\circled{2} Masked attention} In contrast to concatenating an image-level $\texttt{[CLS]}$ token as in \citep{caron2021emerging,zhou2021ibot,oquab2023dinov2}, we add an object-level class token $\texttt{[OBJ]} \in \mathbb{R}^{1 \times D}$ to the input patch sequences $\tilde{x}^{\uupper}, \tilde{x}^{\vupper} \in \mathbb{R}^{HW \times D}$: $[\texttt{[OBJ]}, \tilde{x}] \in \mathbb{R}^{(1+HW) \times D}$. The functionality of the $\texttt{[OBJ]}$ token is to represent only the features of the target object in contrast to 
$\texttt{[CLS]}$ token representing the whole image. 
Using masked attention, the $\texttt{[OBJ]}$ token only attends to those patches where the object is present based on the object binary masks $\tilde{y}^{\uupper}, \tilde{y}^{\vupper}$. In other words, we prevent the $\texttt{[OBJ]}$ token from attending to the patches where the object is not present. Again, we use the scenario where the teacher network takes \colorboxtight{yellowfill}{\texttt{view 1}} as input and the student network takes \colorboxtight{redfill}{\texttt{view 2}}:
\begin{align}
    z^{\uupper}_{\texttt{[OBJ]},t}, z^{\uupper}_{\texttt{patches},t} &= b_t ([\texttt{[OBJ]}^{\uupper}, \tilde{x}^{\uupper}], \texttt{obj-attn-mask}=\tilde{y}^{\uupper}), & \texttt{teacher - \colorboxtight{yellowfill}{view 1}} \\
    z^{\vupper}_{\texttt{[OBJ]},s}, z^{\vupper}_{\texttt{patches},s} &= b_s ([\texttt{[OBJ]}^{\vupper}, \tilde{x}^{\vupper}], \texttt{obj-attn-mask}=\tilde{y}^{\vupper}), & \texttt{student - \colorboxtight{redfill}{view 2}}
\end{align}
Notice that each transformer layer uses the object segmentation mask as input to the \texttt{MaskedMultiHeadAttention} (\texttt{MaskedMHA}) block as in \cref{fig:masked-attention} to update the attention scores of the $\texttt{[OBJ]}$ token. 
This leads to object-level representations that are highly nonlinear mixtures of the corresponding patch tokens, as opposed to works that consider average pooling of the patches \citep{henaff2021efficient,henaff2022object,lebailly2023cribo}.

In standard ViTs, \texttt{[CLS]} token can attend to any other token, including large, textured, or crop-overlapping background patches. These non-informative tokens often steal some attention from the tokens that correspond to important foreground objects. Our masked-attention design breaks this \textit{attention competition} by allowing \texttt{[OBJ]} token to \emph{pool} exclusively from tokens that fall inside the segmentation mask. This simple masking eliminates background ``free-riders'', thereby yielding a cleaner object embedding with a higher signal-to-noise ratio. Importantly, the restriction applies \emph{only} to the \texttt{[OBJ]} token, i.e., the patch tokens belonging to the object still participate in full, unmasked self-attention with the rest of the patches in the image. Thus they can pull in whatever context is genuinely informative. For example, barn walls and grass texture in \cref{fig:IN-multi-obj} carry information about the cow tokens; hence, they may help object tokens to better describe the object. In short, masked attention resolves the attention competition problem at pooling time while preserving the rich cross-token interactions that make transformer features powerful in the first place.

\textbf{Object-level objective.} MLP prediction heads take the representations $z_{\texttt{[OBJ]}}$ as input and produce probability vectors $p_s, p_t$, e.g., $p_{\texttt{[OBJ]},s} = h_s(z_{\texttt{[OBJ]}})$. 
For clarity, we again provide a simplistic example computing the cross-entropy loss only in one direction:
We take cross-entropy loss between probability vectors $p_s, p_t$ that correspond to distinct views $x^{\uupper}, x^{\vupper}$:
\begin{align}
    p_{\texttt{[OBJ]},t}^{\uupper} &= h_t(z^{\uupper}_{\texttt{[OBJ]}}), & \texttt{teacher - \colorboxtight{yellowfill}{view 1} [OBJ]} \\
    p_{\texttt{[OBJ]},s}^{\vupper} &= h_s(z^{\vupper}_{\texttt{[OBJ]}}), & \texttt{student - \colorboxtight{redfill}{view 2} [OBJ]} \\
    \mathcal{L}_{\texttt{[OBJ]}} &= \texttt{CrossEntropy}( p_{\texttt{[OBJ]},t}^{\uupper}, p_{\texttt{[OBJ]},s}^{\vupper}) & \texttt{object-level loss}
\end{align}
while the loss is symmetric: $\mathcal{L}_{\texttt{[OBJ]}} = \frac{1}{2} (\texttt{CE}( p_{\texttt{[OBJ]},t}^{\uupper}, p_{\texttt{[OBJ]},s}^{\vupper}) + \texttt{CE}( p_{\texttt{[OBJ]},t}^{\vupper}, p_{\texttt{[OBJ]},s}^{\uupper}))$.

\textbf{Final loss.} Our final loss sums the object-level loss with the patch-level loss described in \cref{sec:pre}:
\begin{align}
    \mathcal{L}_{\text{ODIS}} = \mathcal{L}_{\texttt{[OBJ]}} + \mathcal{L}_{\texttt{[PATCH]}}.
\end{align}
As the patch-level masking strategy, we use random block masking as in \citep{zhou2021ibot}.

\textbf{Discussion on the use of object segmentation maps} 
Modern SSL adopts weak supervision signals such as paired text, yet it still overlooks the simplest one: the segmentation masks already bundled with ImageNet-1k, COCO, and many other datasets. In ODIS we treat these masks as free supervision, feeding object-aware crops during pre-training for the network to learn spatially grounded features. In case masks are not available at inference time, we propose to run a lightweight class-aware segmentation tool and pool only from the predicted object region. This straightforward tweak lifts accuracy across every benchmark we tried, without extra labels or hyperparameter tuning. Whenever masks are available or can be generated automatically, SSL pipelines should default to using them.

\subsection*{Implementation details}

We follow the ViT architectures and the pretraining setups in previous works \citep{caron2021emerging,zhou2021ibot}, as further detailed in \cref{appsec:impl}. We use ViTs of different sizes, ViT-Small/16, ViT-Base/16 and ViT-Large/16 with patch size equal to $16$.

\textbf{Object segmentation maps.} For COCO and IN1k, we use the provided ground-truth segmentation maps for the main experiments. 
For COCO, each image has on average $\sim7$ distinct object instances of $\sim150$ object classes.
For IN1k, a single object segmentation map is provided for each image, locating the main object. For IN1k, all $50$k validation images have a valid segmentation map, while only $500$k $/1.2$M training images have one. For the images missing the segmentation map, we assume that the main object covers the whole image.
On IN1k, we also ablate different object segmentation maps produced by off-the-shelf segmentation models \citep{redmon2016you,carion2020end}. YOLO \citep{redmon2016you} and a multi-modal ViT \citep[MAVL]{maaz2022class} provide class-agnostic segmentation maps, possibly with multiple distinct objects for each image.

\section{Experiments}

Our main goal is to learn visual representations useful for downstream tasks. First, we choose an image-level representation task: standard self-supervised benchmarking on ImageNet-1k (IN1k)\citep{chen2020simple,caron2021emerging,zhou2021ibot,oquab2023dinov2}, that is, classification using the frozen features with $k$-NN classifier or linear probing (LP). As IN1k images are intended to be object-centric, i.e., contain a single dominant object, this task can also be viewed as an object-level task convenient to assess our object-level representations. Second, we choose a patch-level task to investigate how object-level distillation affects patch-level representations: in-context scene understanding, also referred as dense nearest neighbor retrieval \citep{balazevic2024towards,lebailly2023cribo}.\looseness-1

\subsection{Standard self-supervised benchmark: Classification on IN1k}
\label{subsec:IN-class}

To measure the quality of frozen object representations, we follow the standard self-supervised benchmark on IN1k. We freeze the ViT (teacher) backbone at test time and use the frozen visual features to build a simple classifier. The standard classifiers are $k$-NN and linear probing. We follow the evaluation setups used in DINO \citep{caron2021emerging}, iBOT \citep{zhou2021ibot} and DINOv2 \citep{oquab2023dinov2}, which (i) sweep over $k$ values for the model selection of the $k$-NN classifier and (ii) sweep over learning rates for the model selection of the linear classifier.

\newcommand{\upchange}[1]{\textcolor{green!50!black}{\textcolor{green!50!black}{\scriptsize $\uparrow$ \text{#1}}}}
\begin{table}
\caption{\textbf{$k$-NN and linear probing (LP) accuracy on ImageNet-1k}. `Use Masks' refers to whether the ground-truth ImageNet masks are provided to the model at inference time or not. To obtain ``DINO/iBOT + Masks'' results, we incorporate masked attention into publicly available checkpoints.}
\footnotesize
\label{tab:in1k}
  \centering
  \begin{tabular}{llrclcll}
    \toprule
    Model & Backbone & \#Params & Epochs & Pretrain. & Use Masks & $k$-NN & LP \\
    \midrule
    \textit{ViT-Small} & & & & \\[1.5pt]
    DINO & ViT-S/16 & 21M & 800 & IN1k & \xmark & $74.5$ & $77.0$ \\
    iBOT & ViT-S/16 & 21M & 800 & IN1k & \xmark & $75.2$ & $77.9$ \\
    ODIS & ViT-S/16 & 21M & 800 & IN1k & \xmark & $75.9$ & $78.2$ \\
    DINO+Masks & ViT-S/16 & 21M & 800 & IN1k & \cmark & $75.6$ & $79.0$ \\
    iBOT+Masks & ViT-S/16 & 21M & 800 & IN1k & \cmark & $76.2$ & $80.1$ \\
    \rowcolor{LightCyan}
    ODIS+Masks & ViT-S/16 & 21M & 800 & IN1k & \cmark & $\mathbf{78.5}$ \upchange{3.2} & $\mathbf{81.1}$ \upchange{3.2} \\
    \midrule
    \textit{ViT-Base} \\[1.5pt]
    DINO & ViT-B/16 & 85M & 400 & IN1k & \xmark & $76.1$ & $78.2$ \\
    iBOT & ViT-B/16 & 85M & 400 & IN1k & \xmark & $77.1$ & $79.5$ \\
    ODIS & ViT-B/16 & 85M & 400 & IN1k & \xmark & $78.3$ & $80.5$ \\
    DINO+Masks & ViT-B/16 & 85M & 400 & IN1k & \cmark & $77.6$ & $80.3$ \\
    iBOT+Masks & ViT-B/16 & 85M & 400 & IN1k & \cmark & $78.6$ & $81.6$ \\
    \rowcolor{LightCyan}
    ODIS+Masks & ViT-B/16 & 85M & 400 & IN1k & \cmark & $\mathbf{80.9}$ \upchange{3.8} & $\mathbf{83.2}$ \upchange{3.8} \\
    \midrule
    \textit{ViT-Large} \\[1.5pt]
    iBOT & ViT-L/16 & 307M & 250 & IN1k & \xmark & $78.0$ & $81.0$ \\
    ODIS & ViT-L/16 & 307M & 250 & IN1k & \xmark & $79.6$ & $81.6$ \\
    iBOT+Masks & ViT-L/16 & 307M & 250 & IN1k & \cmark & $79.9$ & $82.5$ \\
    \rowcolor{LightCyan}
    ODIS+Masks & ViT-L/16 & 307M & 250 & IN1k & \cmark & $\mathbf{82.6}$ \upchange{4.6} & $\mathbf{84.6}$ \upchange{3.6} \\
    \midrule
    \textit{DINOv2} \\[1.5pt]
    DINOv2-Dis. & ViT-S/14 & 21M & - & LVD-142M & \xmark & $79.0$ & $81.1$ \\
    DINOv2-Dis. & ViT-B/14 & 85M & - & LVD-142M & \xmark & $82.1$ & $81.4$ \\
    DINOv2-Sc. & ViT-L/14 & 307M & - & IN22k & \xmark & $82.0$ & $84.5$ \\
    DINOv2-Dis. & ViT-L/14 & 307M & - & LVD-142M & \xmark & $83.5$ & $86.3$ \\
    DINOv2-Sc. & ViT-g/14 & 1.1B & - & LVD-142M & \xmark & $83.5$ & $86.5$ \\
    \bottomrule
  \end{tabular}
\end{table}

\begin{wrapfigure}{r}{0.65\textwidth}
    \centering
    \includegraphics[width=0.594\linewidth]{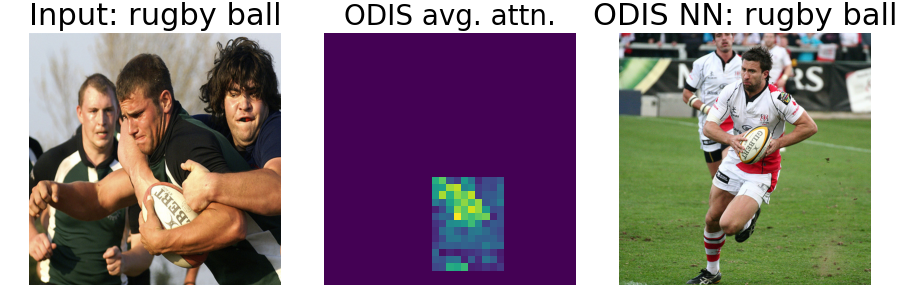} \hspace{-.15cm}
    \includegraphics[trim={8.0cm 0 0 0}, clip, width=.384\linewidth]{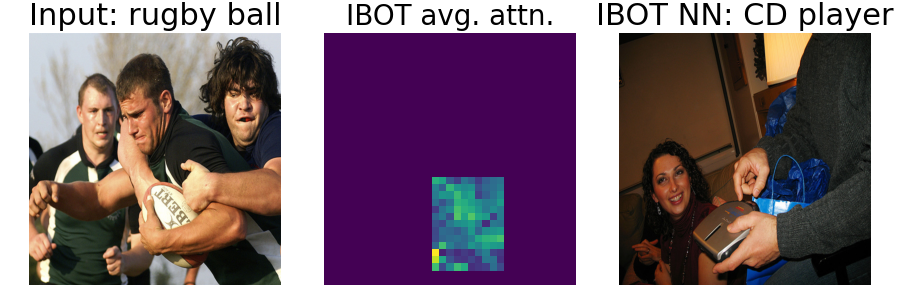}
    \caption{An example input, ODIS and iBOT attention maps using inference-time masks, and retrieved nearest neighbors.
    Despite using the object mask, iBOT mistakenly attends to the hand, while ODIS attends on the correct target object, the rugby ball, demonstrating superior object-level representations.
    }
    \label{fig:ibotfails}
    \vspace{-0.5cm}
\end{wrapfigure}

\textbf{Scenario-1: segmentation masks available during inference} We start with the scenario that models have access to segmentation masks in inference time. Comparing the green ticked rows in \cref{tab:in1k} reveals ODIS clearly outperforms DINO and iBOT. For $k$-NN, the performance gains compared to iBOT are {\color{green!50!black} $+2.3$} for ViT-S, (ii) {\color{green!50!black} $+2.3$} for ViT-B and {\color{green!50!black} $+2.7$} for ViT-L (we note that iBOT improves over DINO by an average of only {\color{green!50!black} $+0.9$}). Please see \cref{fig:ibotfails} for a visual demonstration.

Next, we compare against DINOv2 \citep{oquab2023dinov2}, the current gold standard in self-supervised learning benchmarks. DINOv2 builds on iBOT by introducing ten algorithmic and optimization improvements, curating a high-quality dataset of 142M images, and scaling the model up to 1.1B parameters. Our experiments show that ODIS surpasses the DINOv2 ViT-L model trained on IN22k by {\color{green!50!black} $+0.6$} percentage points in $k$-NN classification accuracy. While it is true that applying segmentation masks at inference would likely improve DINOv2's performance, ODIS is expected to similarly benefit from scaling up the model and data, as well as the same set of algorithmic advances. Since our attempts to fully replicate DINOv2 ViT-L results were unsuccessful, we leave the task of augmenting ODIS with DINOv2-style improvements as a promising direction for future work.

\textbf{Scenario-2: no segmentation masks during inference} Next we turn our attention to this more traditional benchmarking scenario. The red-crossed rows in \cref{tab:in1k} show that ODIS again outperforms DINO and iBOT by a significant margin. It implies that our backbone has learned richer representations that generalize better than DINO and iBOT. We note that larger models benefit more from incorporating our ideas into training.

\textbf{Segmentation masks improve model evaluation by isolating object-specific representations.}
\Cref{fig:app:whyneedmask} presents examples in which nearest neighbor retrieval based on \texttt{[CLS]} token of an iBOT-pretrained ViT fails when the input image is a complicated scene or contains multiple potential target objects. In all cases, the retrieved image is semantically similar but labeled differently, leading to an incorrect match under the IN1k single-label protocol. Last-layer attention maps reveal that the model attends to \emph{multiple} salient objects, highlighting that the embedding captures mixed object semantics. This entanglement undermines retrieval evaluation, especially when multiple plausible objects exist in the scene. To address this, we advocate the use of segmentation masks during inference to isolate individual object representations, enabling more faithful and interpretable evaluation of pretrained models. \looseness-1

\subsection{Self-supervised pretraining on scene-centric data}

\begin{wraptable}{r}{6cm}
\vspace{-0.4cm}
\caption{\textbf{$k$-NN ImageNet-1k for scene-centric pretraining}. All model sizes are ViT-S/16 with 21M parameters. `M.' refer to whether the ground-truth ImageNet masks are provided to the model at inference time or not.}
\label{tab:knn-scene}
\scriptsize
  \centering
  \begin{tabular}{lclcl}
    \toprule
    Model & Epochs & Pretrain. & M. & $k$-NN \\
    \midrule
    DINO & 300 & Coco & \xmark & $36.9$ \\
    CRIBO & 300 & Coco & \xmark & $38.2$ \\
    iBOT & 300 & Coco & \xmark & $41.8$ \\
    iBOT+M. & 300 & Coco & \cmark & $43.9$ \\
    \rowcolor{LightCyan}
    ODIS+M. & 300 & Coco & \cmark & $\mathbf{46.0}$ \upchange{4.2} \\
    \bottomrule
  \end{tabular}
\vspace{-0.cm}
\end{wraptable} 
In this section, we validate our hypothesis that object-level distillation objective better scales to more complex scene-centric datasets such as COCO. We pretrain a ViT-S/16 model with 21M parameters on the COCO dataset (118k images) using DINO, iBOT and ODIS objectives. We freeze the pretrained models and build $k$-NN classifiers on top of frozen features similar to \cref{subsec:IN-class}. We see a similar trend with \cref{subsec:IN-class}: (i) the $k$-NN performance on IN1k ({\color{green!50!black} $+4.2$}) improves significantly compared to iBOT, and (ii) using masks at inference time improves performance for iBOT ({\color{green!50!black} $+1.7$}).

\begin{table}
    \begin{center}
    \scriptsize
    \addtolength{\tabcolsep}{-0.2pt}
    \caption{\textbf{Dense nearest neighbor retrieval task}. We predict in-context segmentation labels and report mIoU. The models are pretrained on a \textit{Scene-centric} dataset, COCO, or an \textit{Object-centric} dataset, IN1k. The models are divided into two groups: (i) \textit{Patch-level} group contains Hummingbird and CRIBO whose objectives primarily focus on increasing cross-image patch-level correspondence, specialized for the dense nearest neighbor retrieval task, (ii) \textit{Higher-level} group contains MAE, DINO, iBOT and ODIS whose objectives focus on image- or object-level representations.
    }
    \label{tab:dense}
    \label{tab:reproduce-cribo}
    \begin{tabular}{lcrccllllllll} 
        \toprule
         &  &  &  &  & \multicolumn{4}{c}{PASCAL VOC} & \multicolumn{4}{c}{ADE 20k} \\ 
         \cmidrule{6-9} \cmidrule{10-13}
        Model & Back. & \#Par. & Pretrain. & Epochs & 1/128 & 1/64 & 1/8 & 1/1 & 1/128 & 1/64 & 1/8 & 1/1 \\ \midrule
        & & & \textit{Scene-c.} & \\
        \textit{Patch-lvl} & & & & \\[1.5pt]
        CRIBO & ViT-S & 21M & Coco & 300 & 39.1 & 44.0 & 52.8 & 58.1 & 10.9 & 12.8 & 18.4 & 23.4 \\
        \arrayrulecolor{black!20}\midrule
        \textit{Higher-lvl} & & & & \\[1.5pt]
        DINO & ViT-S & 21M & Coco & 300 & 16.2 & 18.4 & 25.5 & 31.9 & 6.1 & 6.9 & 9.7 & 13.0 \\
        MAE & ViT-S & 21M & Coco & 300 & 8.5 & 9.3 & 12.2 & 15.9 & 3.7 & 4.1 & 5.4 & 6.8 \\
        iBOT & ViT-S & 21M & Coco & 300 & 37.3 &  39.5 & 47.3 & 54.7 & 10.2 & 12.2 & 16.7 & 21.3 \\
        \rowcolor{LightCyan}
        ODIS & ViT-S & 21M & Coco & 300 & 42.7 & 43.6 & 51.8 & 57.7 \upchange{3.0} & 11.2 & 13.1 & 17.7 & 22.4 \upchange{1.1} \\
        \arrayrulecolor{black}\midrule
        & & & \textit{Object-c.} & \\
        \textit{Patch-lvl} & & & & \\[1.5pt]
        CRIBO & ViT-S & 21M & IN1K & 800 & 52.7 & 59.3 & 69.3 & 73.2 & 13.7 & 16.5 & 23.2 & 28.3 \\
        \arrayrulecolor{black!20}\midrule
        \textit{Higher-lvl} & & & & \\[1.5pt]
        DINO & ViT-S & 21M & IN1K & 800 & 24.5 & 28.7 & 38.7 & 46.1 & 9.4 & 10.6 & 14.6 & 18.4 \\
        iBOT & ViT-S & 21M & IN1K & 800 & 34.6 & 41.1 & 54.7 & 62.1 & 11.9 & 13.9 & 18.8 & 23.1 \\
        \rowcolor{LightCyan}
        ODIS & ViT-S & 21M & IN1K & 800 & 35.5 & 41.6 & 55.6 & 63.3 \upchange{1.2} & 12.1 & 14.2 & 19.3 & 24.1 \upchange{1.0} \\
        \arrayrulecolor{black}\midrule
        \textit{Patch-lvl} & & & & \\[1.5pt]
        Humming. & ViT-B & 85M & IN1K & 300 & 50.5 & 57.2 & - & 70.5 & 11.7 & 15.1 & - & 28.3 \\
        CRIBO & ViT-B & 85M & IN1K & 400 & 50.5 & 60.3 & 70.8 & 74.9 & 13.2 & 16.5 & 23.6 & 30.0 \\
        \arrayrulecolor{black!20}\midrule
        \textit{Higher-lvl} & & & & \\[1.5pt]
        DINO & ViT-B & 85M & IN1K & 400 & 29.2 & 34.7 & 47.2 & 54.9 & 11.1 & 12.6 & 17.6 & 22.0 \\
        MAE & ViT-B & 85M & IN1K & 1600 & 6.0 & 6.5 & 8.9 & 13.8 & 2.7 & 3.0 & 4.0 & 5.3 \\
        iBOT & ViT-B & 85M & IN1K & 400 & 41.1 & 47.4 & 60.6 & 67.8 & 14.8 & 17.1 & 22.9 & 27.4 \\
        \rowcolor{LightCyan}
        ODIS & ViT-B & 85M & IN1K & 400 & 43.1 & 49.7 & 63.1 & 70.0 \upchange{2.2} & 16.2 & 18.8 & 25.1 & 30.0 \upchange{2.6} \\
        \arrayrulecolor{black}\midrule
        iBOT & ViT-L & 307M & IN1K & 250 & 41.1 & 46.7 & 60.8 & 68.6 & 15.8 & 18.3 & 24.4 & 29.0 \\
        \rowcolor{LightCyan}
        ODIS & ViT-L & 307M & IN1K & 250 & 44.6 & 51.2 & 65.4 & 72.6 \upchange{4.0} & 17.1 & 19.7 & 26.1 & 31.0 \upchange{2.0} \\
        \bottomrule
    \end{tabular}
    \end{center}
\end{table}

\subsection{Patch-level task: Dense Nearest Neighbor Retrieval}

We evaluate the usefulness of patch-level representations with the dense nearest neighbor retrieval task \citep{balazevic2024towards}, which extends the standard image-level self-supervised benchmark to patches. Similar to $k$-NN classification, each patch is assigned a label by aggregating labels from a memory bank of reference patches, but here the final prediction uses cross-attention weights rather than a simple distance metric. See \cref{ssec:dense} for the detailed task description and evaluation setup. We report mean Intersection-over-Union (mIoU) on two segmentation benchmarks: PASCAL VOC \citep{everingham2015pascal} and ADE20k \citep{zhou2017scene}, as summarized in \cref{tab:dense}.

\textbf{Results.} We see substantial mIoU performance gains for ODIS patch representations compared to iBOT across all datasets, all subsampling factors and all model sizes. On PASCAL VOC with subsampling factor equal to $1$, the performance gains compared to iBOT are (i) {\color{green!50!black} $+1.2$} for ViT-S, (ii) {\color{green!50!black} $+2.2$} for ViT-B and {\color{green!50!black} $+4.0$} for ViT-L. On ADE20k with subsampling factor equal to $1$, the performance gains compared to iBOT are (i) {\color{green!50!black} $+1.0$} for ViT-S, (ii) {\color{green!50!black} $+2.6$} for ViT-B and {\color{green!50!black} $+2.0$} for ViT-L.

Hummingbird \citep{balazevic2024towards} and CRIBO \citep{lebailly2023cribo} provide the best performance on this task as their learning objectives primarily focus on increasing cross-image patch-level correspondence. However, their patch-level performance comes at the cost of worse image-level representations. In \cref{tab:knn-scene}, we show that IN1k $k$-NN accuracy of CRIBO pretrained on COCO is significantly lower than iBOT and ODIS. In addition, for model size ViT-B on ADE20k, we see that ODIS mIoU is on par with CRIBO mIoU while surpassing Hummingbird mIoU.

\subsection{Ablations}
\begin{wraptable}{r}{5cm}
\vspace{-0.4cm}
\caption{\textbf{Ablation study on different object segmentation maps.}}
\label{tab:knn-masks}
\small
  \centering
  \begin{tabular}{llc}
    \toprule
    Model & Segmenter & $k$-NN \\
    \midrule
    iBOT+M. & - & $76.2$ \\
    \midrule
    ODIS+M. & YOLO & $76.8$ \\
    ODIS+M. & MAVL & $77.1$ \\
    ODIS+M. & Ground-truth & $78.5$ \\
    \bottomrule
  \end{tabular}
\vspace{-0.1cm}
\end{wraptable}
We ablate the loss components, local-crop configurations, object sampling strategies, and off-the-shelf segmentation masking methods (please see \cref{appsec:abl} for details). In summary, we discover \textit{(i)} excluding image-level loss improves our accuracy, \textit{(ii)} local crops are drawn randomly from entire image, \textit{(iii)} sampling larger objects more often yields better results, and \textit{(iv)} using off-the-shelf tools to extract segmentation masks for pretraining still increases kNN accuracy (\cref{tab:knn-masks}). All models are ViT-S/16, pretrained on IN1k for 800 epochs. They use the object masks provided by the corresponding segmenter for pretraining while using the ground-truth object maps at inference time.

\section{Conclusion}

In this work, we explore object-level self-distillation (ODIS) for pretraining vision foundation models. 
We show empirically that ODIS learns general-purpose visual representations that are useful for downstream tasks at both image- and patch-level benchmarks; and it improves downstream task performance significantly over the baseline image-level distillation methods while closing the gap with the large-scale DINOv2 model. 
Our object-level distillation assumes the availability of object segmentation masks, a capability that has become increasingly feasible even for uncurated datasets with modern segmentation models. 
In addition, the network efficiency could be improved if the model distills multiple objects in a single forward pass.
In future work, we plan to scale our method to larger models sizes (e.g., ViT-g) and larger datasets (e.g., IN22k and beyond). 

\bibliography{ms}

\clearpage

\appendix

\section{Extended related work}
\label{appsec:relwork}

\paragraph{Modern segmentation models.}  To enable object-level self-distillation, a model must localize individual objects within images. Whenever ground-truth segmentation maps are available (e.g., in ImageNet \citep{deng2009imagenet} or COCO \citep{lin2014microsoft}), we can directly leverage them. Even in the absence of such annotations, this step is increasingly tractable thanks to modern segmentation models \citep{liu2024grounding,kirillov2023segment,ravi2024sam}, which exhibit robust zero-shot segmentation capabilities on scene-centric datasets \citep{rubinstein2025we}. Harnessing these models for segmentation masks allows us to apply object-level distillation broadly, even in less-curated datasets. \looseness-1

\paragraph{Object-centric learning methods.} Object-centric learning (OCL) approaches \citep{burgess2019monet,locatello2020object,seitzer2022bridging,didolkar2025transfer} also aim to discover object-like structures in images, typically evaluating performance through unsupervised object segmentation. Yet, modern segmentation foundation models \citep{kirillov2023segment} outperform current OCL methods in zero-shot scenarios \citep{rubinstein2025we}, making it uncertain whether OCL is useful for broad vision tasks. Object-centric representations are further assumed to capture compositional structures useful for visual reasoning tasks \citep{ding2021attention,mamaghan2024exploring}, however it remains unclear how transferable their learned object-level features are beyond visual reasoning and segmentation as the quality of their learned representations are not tested on standard benchmarks. 

In contrast, our work focuses on learning object-level representations that prove directly useful in standard self-supervised benchmarks such as ImageNet, e.g., $k$-NN classification. By coupling object-level distillation with segmentation masks, we bridge insights from OCL and large-scale self-supervision, and we anticipate that the resulting representations will also be useful for OCL-related tasks. \looseness-1

\section{Connections with other masked modeling frameworks and graph learning}

\paragraph{Connection to BERT, Masked Image Modeling, Masked Autoencoders}
Masked image modeling with vision transformers draws inspiration from masked language modeling in NLP \citep{devlin2019bert}, where masked words are predicted from their surrounding context. In vision, similar strategies have been applied: models predict masked image patches \citep[Masked Autoencoders]{he2022masked} or discrete visual tokens \citep[BEiT]{bao2021beit} based on neighboring content, leading to highly effective generative frameworks. Self-supervised approaches such as iBOT \citep{zhou2021ibot} and DINOv2 \citep{oquab2023dinov2} extend this idea using masked patch prediction combined with a distillation objective.

Despite their empirical success, these vision models diverge fundamentally from their textual counterparts: while language models predict meaningful and discrete units like words or subword tokens, masked vision models typically predict arbitrary patches, which are often unidentifiable parts of objects or even background. Moreover, whereas text tokenizers increasingly align with linguistic units (syllables or words), vision lacks such semantically grounded units.
In this work, we address this gap by proposing objects, which are the natural semantic units of visual scenes, as prediction targets. Analogous to words in language, objects in images offer coherent, interpretable units for representation learning.

\paragraph{Connection to Graph and Subgraph Pooling}
We can view each image as a fully-connected graph, where nodes represent patches and node representations correspond to patch embeddings. In this view, image-level distillation via \texttt{[CLS]} token corresponds to pooling a graph-level representation from all nodes. This is a hard task to solve. Object-level distillation via \texttt{[OBJ]} token corresponds to pooling a subgraph-level representation where the subgraph is located via segmentation maps. This is a simpler sub-task, that is aligned better with cross-entropy loss for scene-centric images.

\section{Implementation Details}
\label{appsec:impl}

\textbf{ViT.} We follow previous works \citep{caron2021emerging,zhou2021ibot} and use vision transformers \citep{dosovitskiy2020image} in different sizes ViT-Small/16, ViT-Base/16 and ViT-Large/16 as the visual backbone $b(\cdot)$ with patch size equal to $16$. We build on the code base of iBOT \citep{zhou2021ibot}. As commonly done, we use 2 global crops of size $224 \times 224$ with 10 local crops of size $96 \times 96$. The teacher only processes 2 global crops as input, while the student processes all crops. We use shared MLP heads for predicting the image- and patch-level probability vectors, with output dimension $8192$. \looseness-1

\textbf{Pretraining setup.} We pretrain our models on COCO \citep{lin2014microsoft} and ImageNet-1k (IN1k) \citep{deng2009imagenet}. To keep our results comparable, we follow the training setups used for COCO in \citep{lebailly2023cribo} and for IN1k in \citep{zhou2021ibot}. For the COCO dataset, we pretrain ViT-S/16 for 300 epochs. For the IN1k dataset, we pretrain ViT-S/16 for 800 epochs, ViT-B for 400 epochs and ViT-L for 250 epochs. 
We use random block masking that masks $p \sim \mathcal{U}[0.1, 0.5]$ of the patches for the $50\%$ of the global crops \citep{zhou2021ibot}.

\section{Experimental Details}

\subsection{Dense nearest neighbor retrieval}
\label{ssec:dense}

This task extends the standard image-level SSL benchmark to patches \citep{balazevic2024towards}.

\paragraph{Task description.} For the training set, each image is split into $HW$ patches, and patch-label pairs ${(p_i, y_i)}_{i=1}^{HW N_{\text{train}}}$ are recorded, where $y_i$ is obtained by average pooling the pixel labels within patch $p_i$. We encode each patch $p_i$ into a feature vector $k_i = b_t(p_i)$ using the frozen ViT backbone $b_t(\cdot)$, and store a subset of these feature-label pairs in a memory bank $\mathcal{M} = \{(k_i, y_i)\}$ with different subsampling factors $\{1,8,64,128\}$.

At test time, for each query patch $p_j$ in the validation set, we:
\begin{enumerate}
    \item encode $p_j$ to obtain $q_j = b_t(p_j)$,
    \item compute similarities between $q_j$ and all features in $\mathcal{M}$ using cross-attention (softmax-normalized),
    \item predict the patch label $\hat{y}_j$ by a weighted average of the top-$k$ matching labels in $\mathcal{M}$, where each label is weighted by its attention score.
\end{enumerate}
The predicted labels ${\hat{y}_j}$ for all patches of a test image are concatenated and then upsampled to the original image size via bilinear interpolation, yielding a final segmentation map.

\paragraph{Evaluation setup.}
Following \citet{balazevic2024towards,lebailly2023cribo}, we pretrain ODIS and iBOT on both a scene-centric dataset, COCO (118k images), and an object-centric dataset, IN1k ($1.28$M images). We fix the maximum memory bank size $|\mathcal{M}|$ to 10,240,000 and sweep $k \in \{30,50\}$.

\begin{table}
    \begin{center}
    \small
    \caption{\textbf{Effect of pretraining design choices.} We test object representations with $k$-NN on IN1k and test patch representations with mIoU on PASCAL VOC. PMLC: Patch masking for local crops. OALC: Object-aware local cropping. MALC: Masked-attention for local crops using object attention masks. `Use Masks' and `M.' refer to using the object segmentation masks at inference time for $k$-NN classification on IN1k.
    }
    \label{tab:ablation-coco}
    \begin{tabular}{llclccc} 
        \toprule
        Model & Backbone & Epochs & Pretrain. & Use Masks & $k$-NN & mIoU \\ \midrule
        DINO & ViT-S & 300 & COCO & \xmark & 36.9 & 30.5 \\
        iBOT & ViT-S & 300 & COCO & \xmark & 41.8 & 51.0 \\
        iBOT+Masks & ViT-S & 300 & COCO & \cmark & 43.9 & 51.0 \\
        \midrule
        \textit{Loss components} & \\
        ODIS + Masks + $\mathcal{L}_i$ & ViT-S & 300 & COCO & \cmark & 44.5 & 51.0 \\
        ODIS + Masks & ViT-S & 300 & COCO & \cmark & 46.0 & 54.9 \\
        \midrule
        \textit{Local Crop Configuration} & \\
        ODIS+PMLC+OALC+MALC & ViT-S & 300 & COCO & \xmark & 39.1 & 54.8 \\
        +Masks & ViT-S & 300 & COCO & \cmark  & 40.1 & 54.8 \\
        - PMLC & ViT-S & 300 & COCO & \cmark & 41.4 & 54.0 \\
        - OALC & ViT-S & 300 & COCO & \cmark & 42.6 & 54.6 \\
        - MALC (=ODIS+Masks) & ViT-S & 300 & COCO & \cmark & 46.0 & 54.9 \\
        \midrule
        \textit{Object Sampling} & \\
        ODIS+Masks+ random sampl. & ViT-S & 300 & COCO & \cmark & 45.3 & 54.9 \\
        ODIS+Masks+ random area sampl. & ViT-S & 300 & COCO & \cmark & 46.0 & 54.9 \\
        \bottomrule
    \end{tabular}
    \end{center}
\end{table}

\subsection{Computational Resources and Runtime Comparison}
\begin{wraptable}{r}{5cm}
\vspace{-0.4cm}
\caption{\textbf{Runtime comparison.} Pretraining ViT-S on IN1k for 1 epoch with 2 nodes of $4\times$ AMD MI250x (world size of 16).}
\label{tab:runtimes}
\small
  \centering
  \begin{tabular}{lcc}
    \toprule
    Model & Batch size & Time per epoch \\
    \midrule
    DINO & 1024 & 10:28 \\
    iBOT & 1024 & 10:34 \\
    ODIS & 1024 & 15:25 \\
    \bottomrule
  \end{tabular}
\vspace{-0.1cm}
\end{wraptable}
In this section, we report the computational resources used and provide a runtime comparison of our method ODIS with DINO \citep{caron2021emerging} and iBOT \citep{zhou2021ibot}. 
We pretrain all models on ImageNet-1k \citep{deng2009imagenet} for one epoch using a ViT-S backbone. For pretraining ViT-S, we use 2 nodes where each node contains $4\times$ AMD MI250x GPUs. Each GPU has 2 compute dies per resulting in a world size of $2 \times 4 \times 2=16$.
For pretraining ViT-B and ViT-L models, we use 4 and 8 nodes respectively.

ODIS creates a negligible memory overhead, as it only adds an object segmentation mask of shape $H \times W \times 1$ to each global view of size $H \times W \times D$, where $H$ and $W$ are the number of patches along vertical and horizontal axes, and $D$ is the embedding dimension. 
We report a runtime comparison for ViT-S in \cref{tab:runtimes}. 
Although ODIS is currently slower than iBOT during pretraining, we expect performance to improve with future optimization of the object sampling process in data loading, which we leave for future work.

\section{Ablations}
\label{appsec:abl}
Next, we list the findings of our ablation studies. We mainly ablate our method on COCO due to computational constraints, where pretrain a ViT-S for 300 epochs as in \citet{lebailly2023cribo}. We report these results in \cref{tab:ablation-coco}. Additionally, we ablate using external masks for pretraining in IN1k and report the results in \cref{tab:knn-masks}.

\textbf{Loss components.} We experimented with including an auxiliary image-level term $\mathcal{L}_{i}$ or not. Removing it improved patch-level accuracy and left object-level metrics more or less unchanged, so $\mathcal{L}_{i}$ is omitted in the final objective.

\textbf{Local-crop configuration.} The best object representations arise when (i) tokens from local crops attend to all crop patches and (ii) the crops themselves are drawn from general, context-rich regions rather than object-aware windows.

\textbf{Object sampling.} On COCO, sampling objects with probability proportional to their area yields a small but consistent advantage over uniform sampling on object-level evaluations with a similar performance on patch-level evaluations.

\textbf{External masks.} We generate object bounding boxes using two modern segmentation models: YOLO \citep{redmon2016you} and MAVL \citep{maaz2022class}. They are both trained on COCO dataset and provide multi-object, class-agnostic bounding boxes. We sample objects with probabilities proportional to their areas for each forward pass. Even though the training distribution of the segmenter models do not exactly match the target IN1k distribution, using boxes generated by YOLO and DETR raises $k$-NN top-1 accuracy by {\color{green!50!black}+0.4} and {\color{green!50!black}+0.9} respectively compared to the iBOT baseline, reported in \cref{tab:knn-masks}. Yet, the $k$-NN performance further benefits from higher quality ground-truth maps.

\begin{figure}
    \centering
    \includegraphics[width=0.99\linewidth]{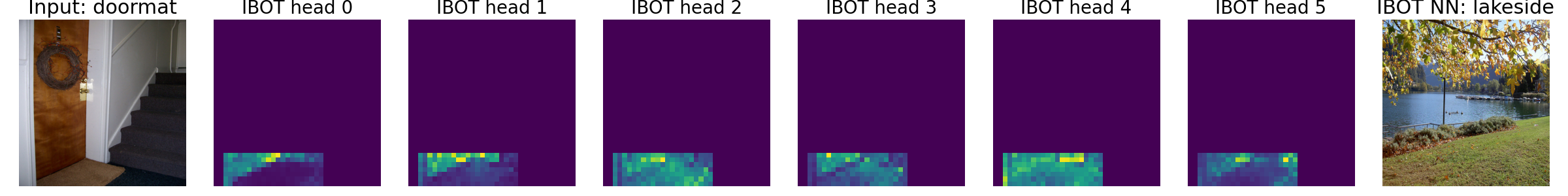}
    \includegraphics[width=0.99\linewidth]{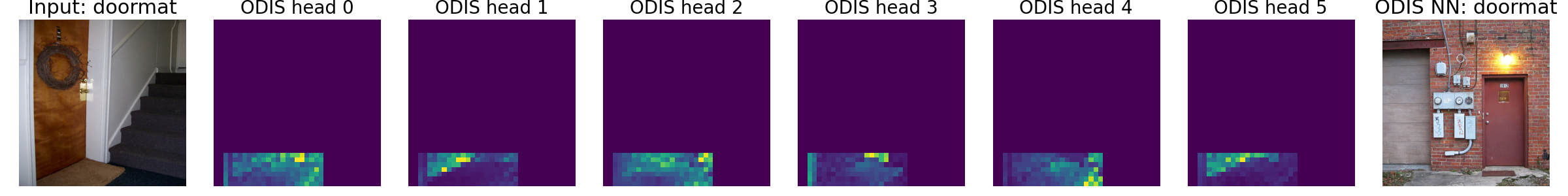}
    \includegraphics[width=0.99\linewidth]{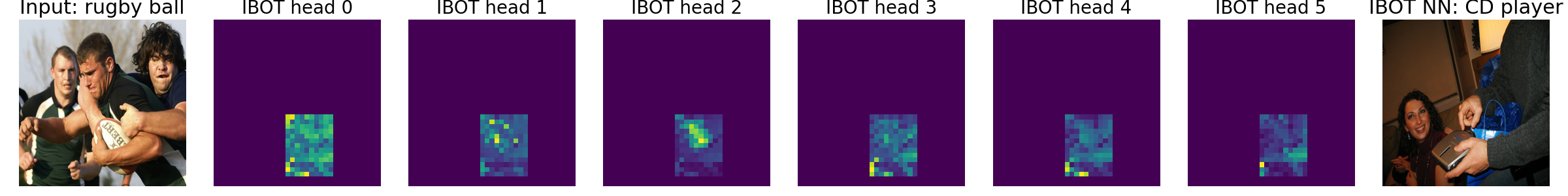}
    \includegraphics[width=0.99\linewidth]{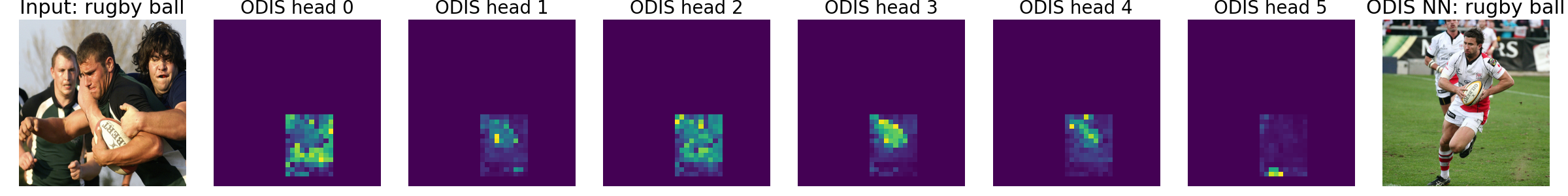}
    \caption{An extended version of \cref{fig:ibotfails}, where all attention heads are visualized.}
    \label{fig:app:ibotfails} 
\end{figure}

\begin{figure}
    \centering
    \includegraphics[width=0.594\linewidth]{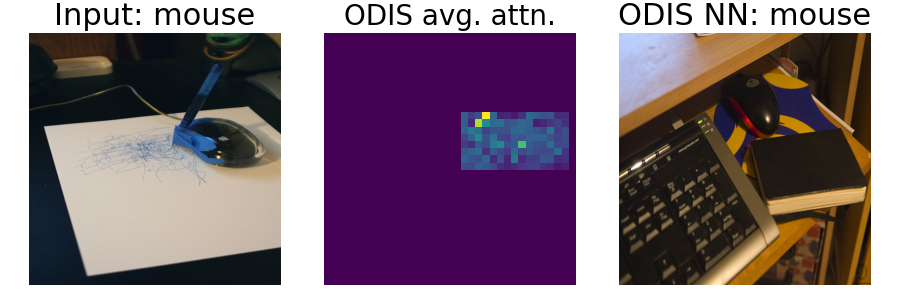} \hspace{-.15cm}
    \includegraphics[trim={8.0cm 0 0 0}, clip, width=.385\linewidth]{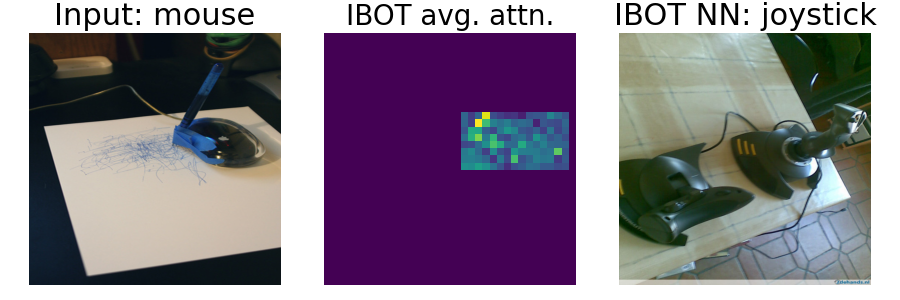}
    \includegraphics[width=0.594\linewidth]{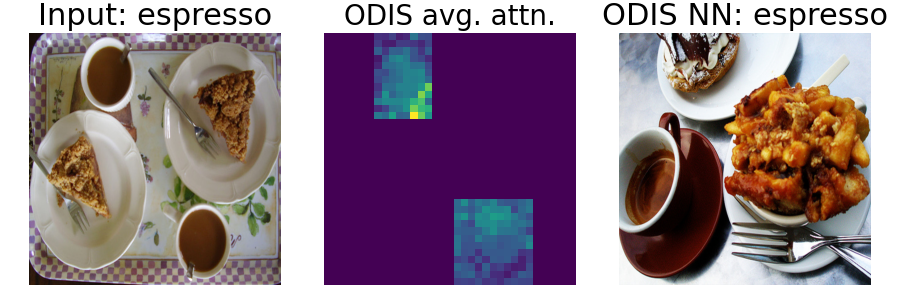} \hspace{-.15cm}
    \includegraphics[trim={8.0cm 0 0 0}, clip, width=.385\linewidth]{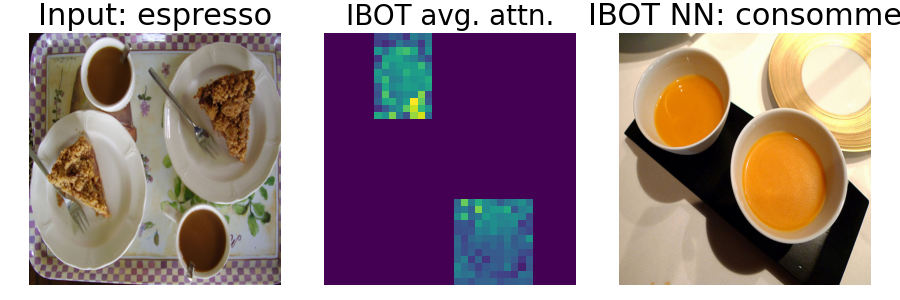}
    \includegraphics[width=0.594\linewidth]{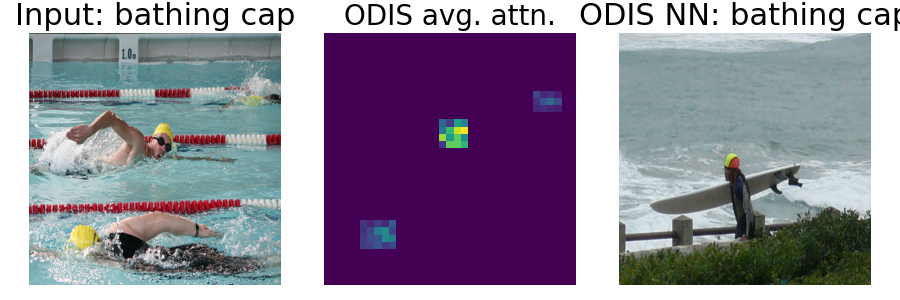} \hspace{-.15cm}
    \includegraphics[trim={8.0cm 0 0 0}, clip, width=.385\linewidth]{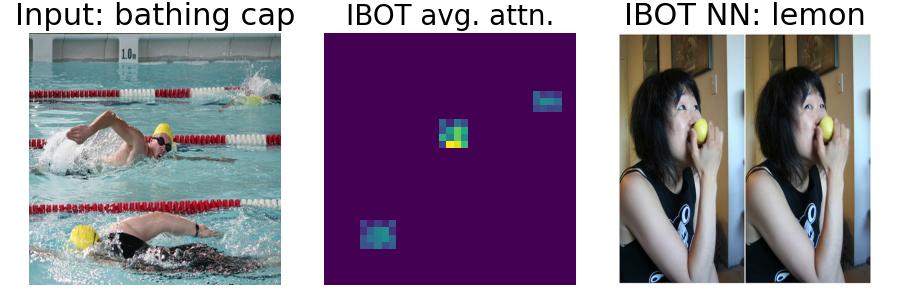}
    \includegraphics[width=0.594\linewidth]{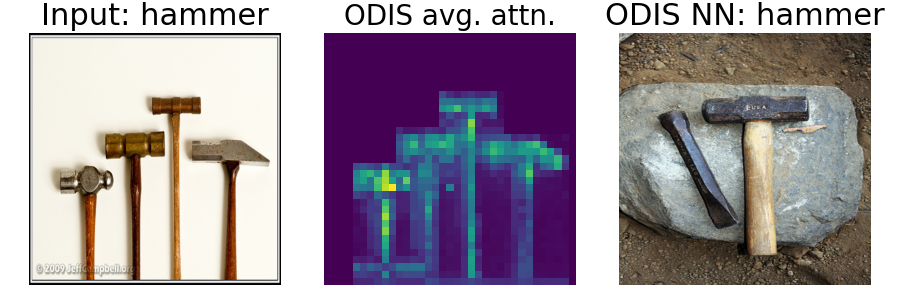} \hspace{-.15cm}
    \includegraphics[trim={8.0cm 0 0 0}, clip, width=.385\linewidth]{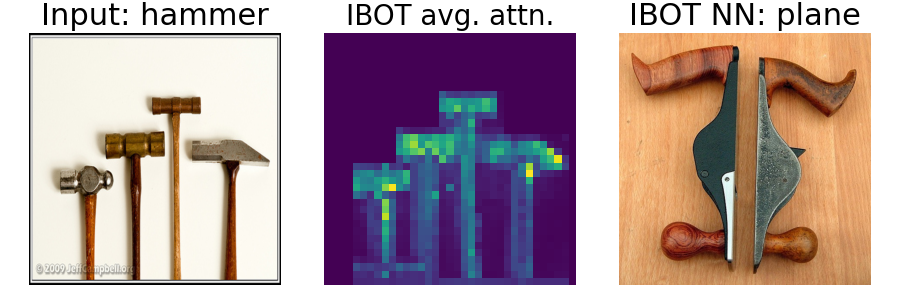}
    \includegraphics[width=0.594\linewidth]{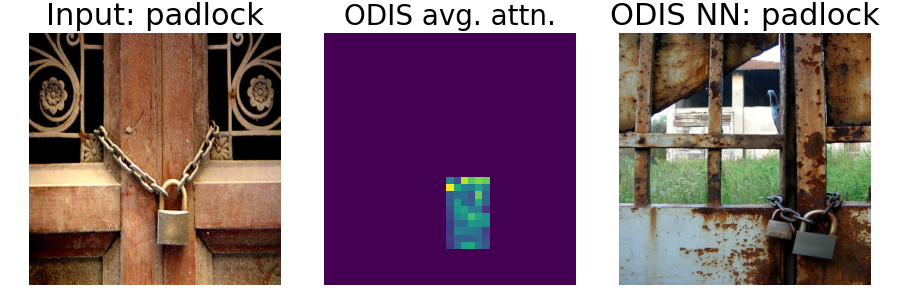} \hspace{-.15cm}
    \includegraphics[trim={8.0cm 0 0 0}, clip, width=.385\linewidth]{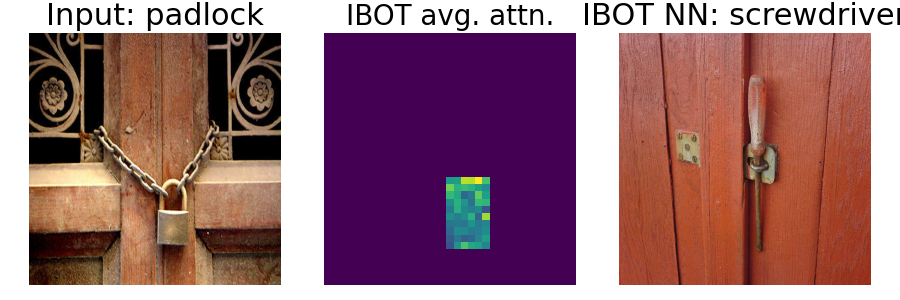}
    \caption{Additional examples showing iBOT's failure despite masked attention.}
    \label{fig:app:ibotfailsmore}
\end{figure}

\begin{figure}
    \centering
    \includegraphics[width=0.46\linewidth]{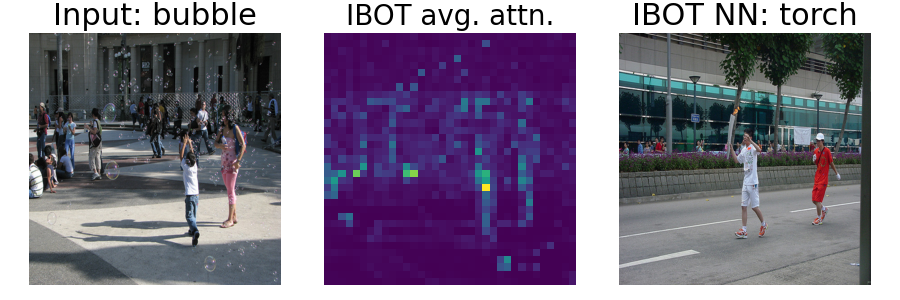} \hspace{.4cm}
    \includegraphics[width=0.46\linewidth]{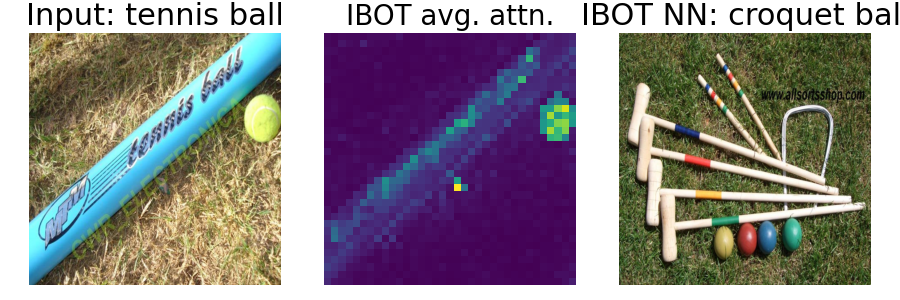}
    \includegraphics[width=0.46\linewidth]{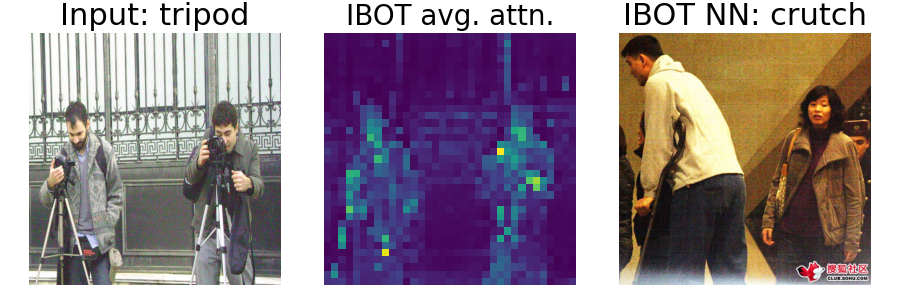} \hspace{.4cm}
    \includegraphics[width=0.46\linewidth]{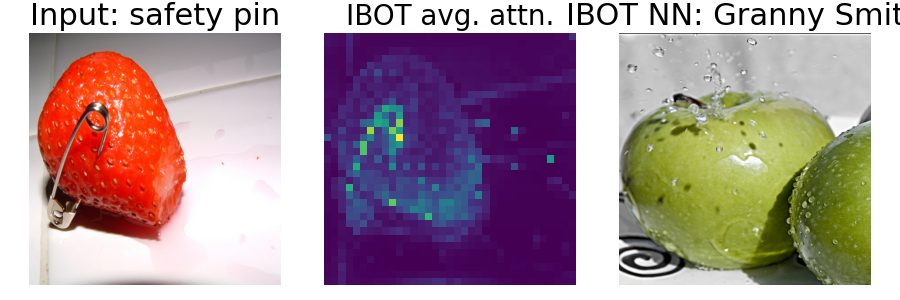}
    \includegraphics[width=0.46\linewidth]{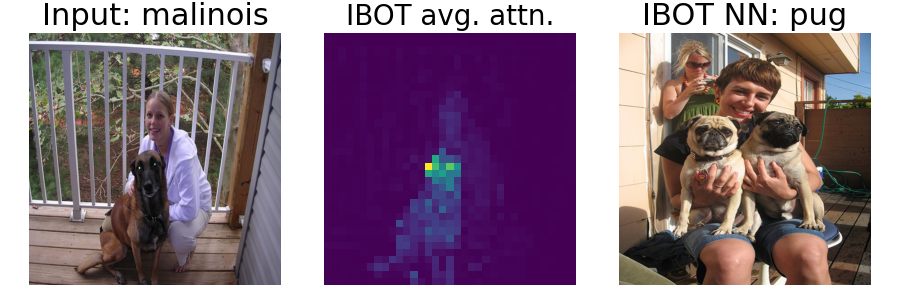} \hspace{.4cm}
    \includegraphics[width=0.46\linewidth]{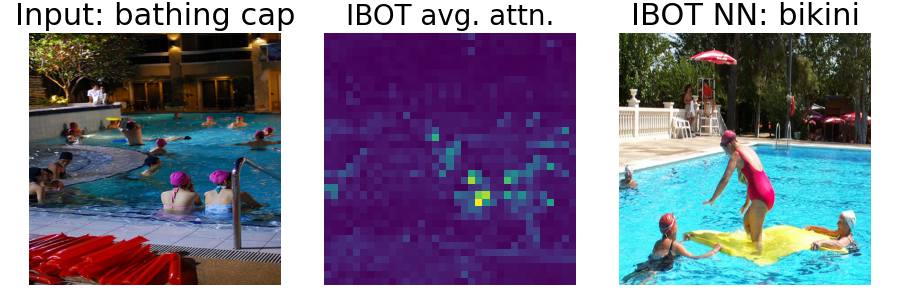}
    \caption{Examples showing how iBOT fails in retrieving a nearest neighbor with the correct class label in the presence of multiple objects. We propose to resolve this by using segmentation masks that specify the target object of interest.}
    \label{fig:app:whyneedmask}
\end{figure}

\end{document}